\pdfoutput=1

\documentclass[11pt]{article}

\usepackage[final]{acl}

\usepackage{times}
\usepackage{latexsym}

\usepackage[T1]{fontenc}

\usepackage[utf8]{inputenc}

\usepackage{microtype}

\usepackage{inconsolata}

\usepackage{graphicx}
\usepackage{multirow}
\usepackage{pifont}
\usepackage{colortbl}
\definecolor{White}{HTML}{ffffff}
\definecolor{NiceRed}{HTML}{5867F5}
\colorlet{c35}{NiceRed!0!White}
\colorlet{c36}{NiceRed!2!White}
\colorlet{c37}{NiceRed!4!White}
\colorlet{c38}{NiceRed!6!White}
\colorlet{c39}{NiceRed!8!White}
\colorlet{c40}{NiceRed!11!White}
\colorlet{c41}{NiceRed!13!White}
\colorlet{c42}{NiceRed!15!White}
\colorlet{c43}{NiceRed!17!White}
\colorlet{c44}{NiceRed!20!White}
\colorlet{c45}{NiceRed!22!White}
\colorlet{c46}{NiceRed!24!White}
\colorlet{c47}{NiceRed!26!White}
\colorlet{c48}{NiceRed!28!White}
\colorlet{c49}{NiceRed!31!White}
\colorlet{c50}{NiceRed!33!White}
\colorlet{c51}{NiceRed!35!White}
\colorlet{c52}{NiceRed!37!White}
\colorlet{c53}{NiceRed!40!White}
\colorlet{c54}{NiceRed!42!White}
\colorlet{c55}{NiceRed!44!White}
\colorlet{c56}{NiceRed!46!White}
\colorlet{c57}{NiceRed!48!White}
\colorlet{c58}{NiceRed!51!White}
\colorlet{c59}{NiceRed!53!White}
\colorlet{c60}{NiceRed!55!White}
\colorlet{c61}{NiceRed!57!White}
\colorlet{c62}{NiceRed!60!White}
\colorlet{c63}{NiceRed!62!White}
\colorlet{c64}{NiceRed!64!White}
\colorlet{c65}{NiceRed!66!White}
\colorlet{c66}{NiceRed!68!White}
\colorlet{c67}{NiceRed!71!White}
\colorlet{c68}{NiceRed!73!White}
\colorlet{c69}{NiceRed!75!White}
\colorlet{c70}{NiceRed!77!White}
\colorlet{c71}{NiceRed!80!White}
\colorlet{c72}{NiceRed!82!White}
\colorlet{c73}{NiceRed!84!White}
\colorlet{c74}{NiceRed!86!White}
\colorlet{c75}{NiceRed!88!White}
\colorlet{c76}{NiceRed!91!White}
\colorlet{c77}{NiceRed!93!White}
\colorlet{c78}{NiceRed!95!White}
\colorlet{c79}{NiceRed!97!White}
\colorlet{c80}{NiceRed!100!White}

\usepackage{booktabs}
\usepackage{enumitem}
\usepackage{amsmath}
\usepackage{subcaption} 
\usepackage{bm}
\usepackage{makecell}

\title{Identifying Fine-grained Forms of Populism in Political Discourse:\\A Case Study on Donald Trump's Presidential Campaigns}

\author{Ilias Chalkidis$^{\alpha\;}$ \quad Stephanie Brandl$^{\beta\;}$ \quad Paris Aslanidis$^{\gamma\;}$\\
$^{\alpha\;}$Department of Computer Science, University of Copenhagen, Denmark \\
$^{\beta\;}$Copenhagen Center for Social Data Science, University of Copenhagen, Denmark \\
$^{\gamma\;}$National Centre for Social Research (EKKE), Greece \\
\small{\texttt{ilias.chalkidis[at]di.ku.dk} \quad \texttt{stephanie.brandl[at]sodas.ku.dk} \quad \texttt{paslanidis[at]ekke.gr}}
}

\begin{document}
\maketitle
\begin{abstract}
Large Language Models (LLMs) have demonstrated remarkable capabilities across a wide range of instruction-following tasks, yet their grasp of nuanced social science concepts remains underexplored. This paper examines whether LLMs can identify and classify fine-grained forms of \emph{populism}, a complex and contested concept in both academic and media debates. To this end, we curate and release novel datasets specifically designed to capture populist discourse. We evaluate a range of pre-trained (large) language models, both open-weight and proprietary, across multiple prompting paradigms. Our analysis reveals notable variation in performance, highlighting the limitations of LLMs in detecting populist discourse. We find that a fine-tuned RoBERTa classifier vastly outperforms all new-era instruction-tuned LLMs, unless fine-tuned. Additionally, we apply our best-performing model to analyze campaign speeches by Donald Trump, extracting valuable insights into his strategic use of populist rhetoric. Finally, we assess the generalizability of these models by benchmarking them on campaign speeches by European politicians, offering a lens into cross-context transferability in political discourse analysis. In this setting, we find that instruction-tuned LLMs exhibit greater robustness on out-of-domain data.
\end{abstract}

\section{Introduction}
\label{sec:intro}

Large Language Models (LLMs)~\cite{openai2023gpt4,llama3,gemini} exhibit unprecedented capabilities, excelling in a plethora of instruction-following tasks such as open-domain question answering, solving arithmetic and coding problems, engaging in creative writing, and more~\cite{bommasani2023holistic,chiang2024chatbot}. 

\begin{figure}[t]
    \centering
    \includegraphics[width=\linewidth]{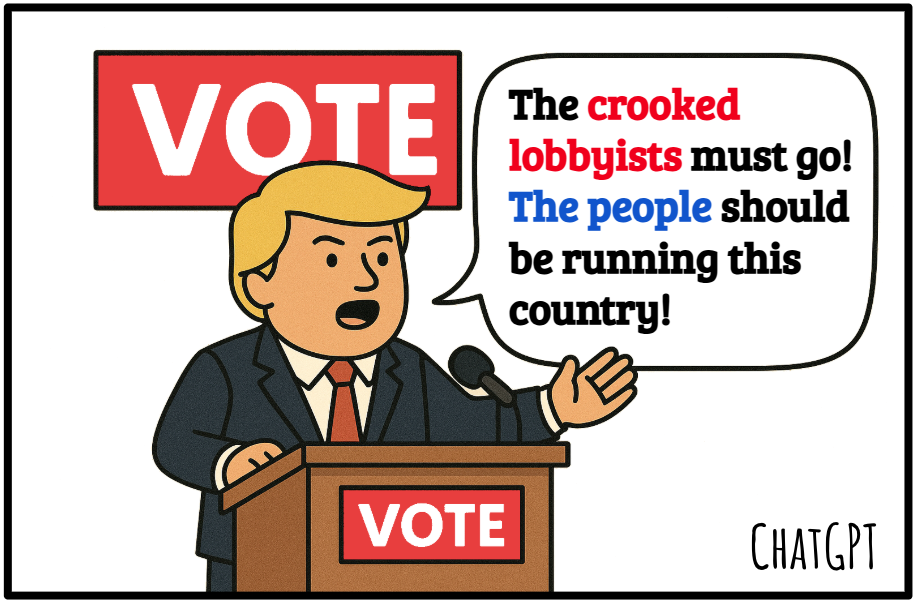}
    \caption{Example of populist political discourse under our working definition, with \textcolor{red}{\textbf{negative}} invocations towards \emph{elites}, and \textcolor{blue}{\textbf{positive}} ones towards the \emph{people}.} 
    \vspace{-1mm}
    \label{fig:demo}
    \vspace{-1mm}
\end{figure}

However, we know little about how well LLMs perform on tasks involving social science concepts that researchers use to investigate sociopolitical phenomena. In this paper, we explore whether LLMs can adequately understand the concept of \emph{populism} (Figure~\ref{fig:demo}) and detect its presence in political discourse. Our goal is to enable the analysis of large-scale corpora and extract valuable insights into how politicians employ populist language.

\begin{figure*}
    \centering
    \includegraphics[width=0.9\textwidth]{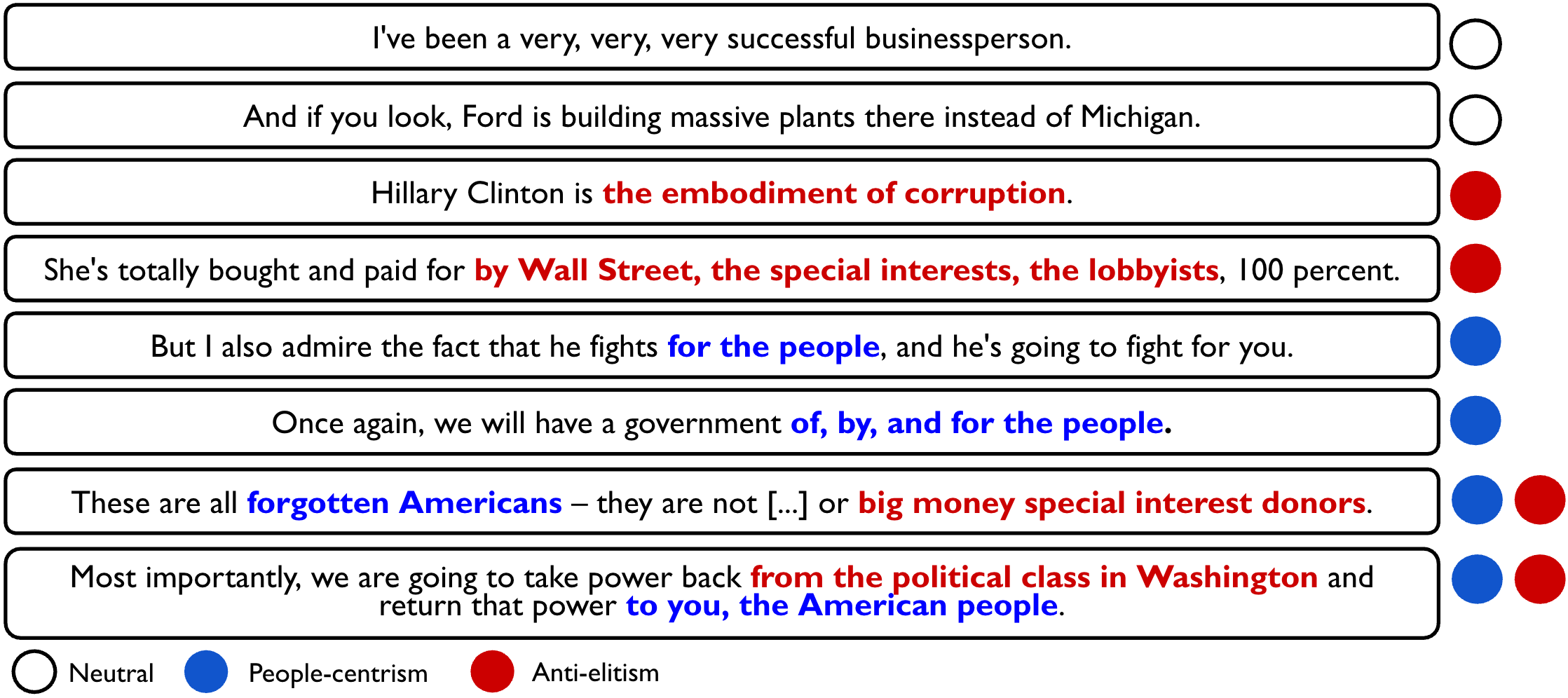}
    \caption{Color-coded examples from the \textsc{Trump-2016} dataset (Section~\ref{sec:datasets}) with two examples per class (neutral, people-centrism, anti-elitism, fully populist). Relevant parts of the sentence are highlighted in red or blue.}
    \label{fig:examples}
    \vspace{-2mm}
\end{figure*}

\paragraph{What is populism?}
\emph{Populism} has attracted growing scholarly attention in recent decades, yet it remains a contested concept, with multiple theoretical traditions offering distinct analytical perspectives~\cite{stavrakakis2024research}. Populism encompasses a broad range of political expressions, spanning both left- and right-wing movements and including authoritarian and democratizing projects. On the political right, figures such as Donald Trump, Jair Bolsonaro, Viktor Orbán, and Marine Le Pen are routinely characterized as populists; on the left, the label has been applied to politicians like Bernie Sanders, Alexandria Ocasio-Cortez, Jeremy Corbyn, and Jean-Luc Mélenchon. Owing to its prominence in political debates, \textit{populism} has become a highly contested label, often equated with \textit{extremism} and \textit{anti-democratic} views, or reduced to simple \textit{demagogy}.

The academic literature on the topic broadly divides into three major schools of thought:\vspace{1mm}

\noindent\hspace{1mm} (a) \emph{Top-down manipulation theories} conceptualize populism as a power-seeking strategy employed by non-ideocratic, opportunistic leaders who bypass institutional mediation to establish direct, quasi-personal relationships with unorganized mass constituencies~\cite{weyland_2001,weyland_2017}; \vspace{1mm}

\noindent\hspace{1mm} (b) \emph{Ideational approaches} define populism as a thin-centered ideology that attaches itself to core ideologies (e.g.~socialism, conservatism) and frequently undermines democratic institutions through its anti-establishment stance and rejection of standard liberal democratic norms \cite{Mudde_2004,Mudde_2007, mudde2017populism, hawkins2018ideational}; \vspace{1mm}

\noindent\hspace{1mm}  (c) \emph{Discursive approaches} define populism as a form of political discourse that constructs a fundamental antagonism between "the people" and "the elites" in order to challenge the status quo~\cite{laclau_2005,stavrakakis-2024,aslanidis-2024}.\vspace{1mm} 

The discursive school of thought aligns with understanding populism as a collective action frame, i.e. a strategic interpretive schema employed by political entrepreneurs to unify diverse societal grievances into a single narrative that champions popular sovereignty over elite domination ~\cite{aslanidis_2016,aslanidis-2024}. This constructivist lens emphasizes the role of meaning-making in sociopolitical mobilization and highlights how populist actors weave together seemingly disparate demands under an ``us'' versus ``them'' (specifically, "people" versus "elites") rhetoric. This approach is particularly well-suited to the computational detection of populism, given its reliance on specific discursive patterns and lexical choices. 

\paragraph{Working Definition}
In this study, we adopt the discursive approach, defining populism as a rhetorical device (a collective action frame) that invokes the principle of popular sovereignty to accentuate the notion of the common ``people'' and their opposition to self-serving elite groups that manipulate the system for their own benefit (Figures~\ref{fig:demo}-\ref{fig:examples}). Accordingly, we identify two core components in populist discourse: (i) people-centrism, i.e., positive references to the "people", and (ii) anti-elitism, i.e., negative references to "the elites".

\paragraph{Scope of our Work}
The goal of this paper is to address the challenge of detecting nuanced manifestations of populism in large-scale textual corpora, using contemporary Natural Language Processing (NLP) techniques. By reducing the limitations of manual annotation and minimizing observational bias, our approach allows for the systematic identification of fine-grained populist patterns in political discourse. As a case study, we analyze Donald Trump’s political rhetoric from 2015 to 2025, focusing on his campaign speeches during the 2016, 2020, and 2024 U.S. presidential elections. More broadly, we seek to provide a methodology that can be applied to other social concepts and use cases across social science disciplines.

\paragraph{Contributions}

We curate and release 3 new datasets (Section~\ref{sec:datasets}), which we leverage to train and benchmark a range of pre-trained (large) language models, including both open-weight and proprietary models, in various settings (Section~\ref{sec:benchmarking}-\ref{sec:llm_analysis}). We then apply our best-performing model to automatically classify and analyze hundreds of Donald Trump's speeches over a decade, gaining valuable insights into his strategic deployment of populist discourse (Section~\ref{sec:trump_analysis}). Finally, we test the generalizability of the best-performing models by benchmarking them on campaign speeches by European politicians, examining their capacity for transfer learning on out-of-domain data and identifying potential performance discrepancies (Section~\ref{sec:eu_benchmarking}).

\section{Task \& Datasets}

\subsection{Task Definition}
\label{sec:task_def}

The goal of this task is to develop a method for identifying fine-grained forms of populist discourse. As outlined in our working definition, populism has two main components: \emph{people-centrism} and \emph{anti-elitism}, which need to operate in tandem before a discursive unit, e.g., a political speech or social media post, can be labeled as populist to some degree. To enable a fine-grained analysis of political content and to leverage high-frequency data for examining variations in populist discourse, we move beyond full speeches or paragraphs as the unit of analysis and instead focus on a lower-level discursive structure: the sentence. We frame the task as a three-way multi-label classification, in which each sentence may be labeled as \textit{neutral}, \textit{anti-elitist}, or \textit{people-centric}. The co-occurrence of the latter two labels designates a \textit{fully populist} sentence (see the last two examples in Figure~\ref{fig:examples}).

\begin{table}[h]
    \centering
    \resizebox{\columnwidth}{!}{
    \begin{tabular}{c|r|r|c}
         \textbf{Dataset Name} & \textbf{\#S} & \textbf{\#I} & \textbf{Labeled}\\
         \midrule
         \textsc{Trump-2016} & 70 & $\sim$15K & \ding{52}\\
         \textsc{Trump-Chronos} & 713& $\sim$656K& \ding{56} \\
         \textsc{EU-OOD}  & 6 & $\sim$1.8K & \ding{52}\\
    \end{tabular}
    }
    \caption{Datasets released as part of our study; \#S refers to number of speeches, and \#I refers to numbers of instances (sentences).}
    \label{tab:datasets}
    \vspace{-2mm}
\end{table}

\subsection{Datasets}
\label{sec:datasets}

As part of this study, we curate and release three new datasets for the identification of fine-grained forms of populist discourse, as described in Table~\ref{tab:datasets}.

\paragraph{Donald Trump (2016)}

We curate and release a dataset of 70 presidential campaign speeches by Donald Trump during the Republican primaries and the 2016 general election campaign, dubbed \textsc{Trump-2016}. The speech transcripts, collected from UC Santa Barbara’s American Presidency Project, span from June 2016 to January 2017, and constitute a mix of actual speech transcripts and prepared remarks released by the Trump campaign.\footnote{\url{https://www.presidency.ucsb.edu/}} Transcripts were curated manually to correct transcription errors, remove chants and other audience interventions, and eliminate remarks by guest speakers or third parties. We also include as metadata the \emph{date} and \emph{location} of each speech.

The speeches were subsequently segmented into sentences, which were annotated according to the 3-class labeling scheme defined in Section~\ref{sec:task_def} by four undergraduate students from Yale University's Department of Political Science, under the supervision of the project's domain expert.\footnote{See Appendix~\ref{sec:guidelines} for details on the annotation process.} 
We use this dataset for training and evaluation purposes, splitting it chronologically into training (56 speeches) and test (14 speeches) subsets.

\begin{table}[h]
    \centering
   \resizebox{\columnwidth}{!}{
    \begin{tabular}{cc|rr}
         \multicolumn{2}{c|}{\textbf{Category}} & \multicolumn{2}{c}{\textbf{\#Instances}}  \\
         \midrule
         Neutral & (N) & 13,910 & (92.6\%) \\
         Anti-elitism & (AE) & 826 & (5.5\%) \\
         People-centrism & (PC) & 517 & (3.4\%) \\
    \end{tabular}
    }
    \caption{Distribution of the three categories in the \textsc{Trump-2016} dataset. Note: \textit{fully populist} sentences are labeled with both populist categories (AE, PC). }
    \label{tab:trump_dist}
    \vspace{-2mm}
\end{table}

Table~\ref{tab:trump_dist} presents the label distribution. Under our strict definition and when annotating at the sentence level, populist invocations appear relatively sparse in political discourse, even in the case of a characteristically populist politician. We emphasize that computational methods should be complemented by qualitative analysis to assess whether these sparse populist elements nonetheless convey the dominant tone of the overall speech.

\paragraph{Donald Trump (2015-2025)}
We also curate and release a much larger collection of 713 unannotated speeches by Donald Trump, spanning a decade from June 2015 to April 2025. Dubbed \textsc{Trump-Chronos}, the dataset comprises a total of 656,136 sentences. This additional corpus is used to analyze historical trends in Trump's populist discourse across multiple election cycles. Having compiled a comprehensive list of Donald Trump's political rallies in this period, we sourced the speech transcripts from UC Santa Barbara's American Presidency Project. For a large number of speeches without readily available transcripts, we used the publicly available version of OpenAI's Whisper~\cite{pmlr-v202-radford23a} to auto-transcribe speech audio extracted from YouTube, C-SPAN and other online platforms hosting the relevant video. All transcripts were manually curated, following the same procedure as for \textsc{Trump-2016}. As we aim to capture discourse that Donald Trump communicated directly to crowds, the dataset thus reflects the actual delivered speeches, rather than prepared remarks distributed to the press by campaign managers.\footnote{Accordingly, our dataset excludes interviews, press conferences, and town hall events in which Trump responds solely to open-ended questions from hosts, moderators, or audience members. However, we include town halls that begin with a speech by Trump, omitting the subsequent Q\&A portion. We also exclude formal events such as presidential inaugurations, joint addresses to Congress, and State of the Union speeches, as well as remarks delivered at invitation-only venues, e.g., think tanks, trade unions, fundraisers, or galas, which tend to focus on specific policy issues rather than broad public appeals. By contrast, we retain speeches delivered at GOP conventions (both regional and national) and at annual or semi-annual CPAC events, given their inclusive and rally-like character.}  Metadata for this corpus includes the \textit{date}, \textit{location}, and \textit{campaign period} of each speech.

\paragraph{European Campaign Speeches}

Additionally, we curate (clean and annotate) a smaller dataset of six campaign speeches by five European leaders. Dubbed \textsc{EU-OOD}, it contains transcripts by Marine Le Pen (Rassemblement National, France), Alice Weidel (AfD, Germany), Herbert Kickl (FPÖ, Austria), Éric Zemmour (Reconquête, France), and Alexis Tsipras (SYRIZA, Greece). The first four are identified as right-wing populists, while Tsipras is categorized as a left-wing populist. We use these speeches as Out-of-Domain (OOD) data to evaluate the robustness of the examined models. Since the speeches are in the speakers' native language (French, German, and Greek), we use the DeepL API\footnote{\url{https://www.deepl.com/en/pro-api}} to produce high-quality English translations.\vspace{2mm}

\noindent All newly developed resources (datasets and models) are released to support the replication of our results and to facilitate further experimentation with alternative methods or corpora.\footnote{\url{https://huggingface.co/collections/coastalcph/populism-686e538a71643fb7529d5639}} Our experimental code is also available on GitHub.\footnote{\url{https://github.com/coastalcph/populism-llms}}

\begin{table*}[t]
    \centering
    \resizebox{0.95\textwidth}{!}{
    \begin{tabular}{c|l|c|c|c|c|c|c}
         \textbf{Group} & \textbf{Model Name} & \textbf{Size} & \textbf{FT} & \textbf{Neutral} & \textbf{Anti-Elitism} & \textbf{People-centrism} & \textbf{Macro F1} \\
         \midrule
         \multirow{2}{*}{\textsc{Baseline}} & Dist. Random & - & - & .928 \small{$\pm$}\hspace{3mm}\small{-}\hspace{2mm} & .068 \small{$\pm$}\hspace{3mm}\small{-}\hspace{2mm} & .053 \small{$\pm$}\hspace{3mm}\small{-}\hspace{2mm} & \cellcolor{c38} .350 \small{$\pm$}\hspace{3mm}\small{-}\hspace{2mm} \\
         & TFIDF-SVM & 30K & \ding{52} & .974 \small{$\pm$}\hspace{3mm}\small{-}\hspace{2mm} & .531 \small{$\pm$}\hspace{3mm}\small{-}\hspace{2mm} & .385 \small{$\pm$}\hspace{3mm}\small{-}\hspace{2mm} & \cellcolor{c63} .630 \small{$\pm$}\hspace{3mm}\small{-}\hspace{2mm} \\
         \midrule
         \multirow{2}{*}{\textsc{\makecell[c]{Pre-trained\\ LMs}}} & RoBERTa & 355M & \ding{52} & \textbf{.976} \small{$\pm$}\hspace{3mm}\small{-}\hspace{2mm} & \underline{.661} \small{$\pm$}\hspace{3mm}\small{-}\hspace{2mm} & \textbf{.602} \small{$\pm$}\hspace{3mm}\small{-}\hspace{2mm} & \cellcolor{c75} \textbf{.746} \small{$\pm$}\hspace{3mm}\small{-}\hspace{2mm} \\
         & DeBERTa & 418M & \ding{52} & .971 \small{$\pm$}\hspace{3mm}\small{-}\hspace{2mm} & .639 \small{$\pm$}\hspace{3mm}\small{-}\hspace{2mm} & \underline{.583} \small{$\pm$}\hspace{3mm}\small{-}\hspace{2mm} & \cellcolor{c73} .731 \small{$\pm$}\hspace{3mm}\small{-}\hspace{2mm} \\
         \midrule
         \multirow{11}{*}{\textsc{\makecell[c]{Open-weight\\ LLMs}}} & \multirow{3}{*}{Llama 3.1} &  \multirow{2}{*}{8B} & \ding{56} & .814 \small{$\pm$ .011}  & .320 \small{$\pm$ .005} & .245 \small{$\pm$ .009} & \cellcolor{c44} .460 \small{$\pm$ .008} \\
         & &  & \ding{52} & .965 \small{$\pm$}\hspace{3mm}\small{-}\hspace{2mm} & .646 \small{$\pm$}\hspace{3mm}\small{-}\hspace{2mm} & .497 \small{$\pm$}\hspace{3mm}\small{-}\hspace{2mm} & \cellcolor{c71} .703 \small{$\pm$}\hspace{3mm}\small{-}\hspace{2mm} \\
          & & 70B & \ding{56} & .850 \small{$\pm$ .010} & .439 \small{$\pm$ .009} & .231 \small{$\pm$ .006} & \cellcolor{c51} .507 \small{$\pm$ .008} \\
         \cmidrule{2-8}
          & \multirow{2}{*}{Gemma 3} & 12B  & \ding{56} & .813 \small{$\pm$ .008} & .318 \small{$\pm$ .002} & .220 \small{$\pm$ .009} & \cellcolor{c44} .437 \small{$\pm$}\hspace{3mm}\small{-}\hspace{2mm} \\
          & & 27B & \ding{56} & .819 \small{$\pm$ .009} & .365 \small{$\pm$ .007} & .232 \small{$\pm$ .006} & \cellcolor{c47} .472 \small{$\pm$ .007} \\
          \cmidrule{2-8}
          & \multirow{5}{*}{Qwen 3} & \multirow{2}{*}{8B}  & \ding{56} & .864 \small{$\pm$ .004} & .368 \small{$\pm$ .002} & .286 \small{$\pm$ .005} & \cellcolor{c51} .506 \small{$\pm$ .004} \\
          & & & \ding{52} & .967 \small{$\pm$}\hspace{3mm}\small{-}\hspace{2mm} & .621 \small{$\pm$}\hspace{3mm}\small{-}\hspace{2mm} & .553 \small{$\pm$}\hspace{3mm}\small{-}\hspace{2mm} & \cellcolor{c71} .714 \small{$\pm$}\hspace{3mm}\small{-}\hspace{2mm} \\
          & & \multirow{2}{*}{14B}  & \ding{56} & .916 \small{$\pm$ .001} & .491 \small{$\pm$ .000} & .340 \small{$\pm$ .005} & \cellcolor{c59} .582 \small{$\pm$ .002} \\
          & & & \ding{52} & \underline{.972} \small{$\pm$}\hspace{3mm}\small{-}\hspace{2mm} & \textbf{.674} \small{$\pm$}\hspace{3mm}\small{-}\hspace{2mm} & .575 \small{$\pm$}\hspace{3mm}\small{-}\hspace{2mm} & \cellcolor{c74} \underline{.740} \small{$\pm$}\hspace{3mm}\small{-}\hspace{2mm} \\
          & & 32B & \ding{56} & .875 \small{$\pm$ .001} & .458 \small{$\pm$ .000} & .292 \small{$\pm$ .001} & \cellcolor{c54} .542 \small{$\pm$ .000} \\
          \midrule
         \multirow{2}{*}{\textsc{\makecell[c]{Proprietary\\ LLMs}}} & GPT 4.1 & n/a & \ding{56} & .850 \small{$\pm$} \hspace{3mm}\small{-}\hspace{2mm} & .461 \small{$\pm$} \hspace{3mm}\small{-}\hspace{2mm} & .201 \small{$\pm$} \hspace{3mm}\small{-}\hspace{2mm} & \cellcolor{c50} .503 \small{$\pm$} \hspace{3mm}\small{-}\hspace{2mm} \\
         & Gemini 2.5 Flash & n/a & \ding{56}  & .903 \small{$\pm$} \hspace{3mm}\small{-}\hspace{2mm} & .513 \small{$\pm$} \hspace{3mm}\small{-}\hspace{2mm} & .306 \small{$\pm$} \hspace{3mm}\small{-}\hspace{2mm} & \cellcolor{c57} .574 \small{$\pm$} \hspace{3mm}\small{-}\hspace{2mm} \\
         \midrule
    \end{tabular}
    }
    \vspace{-2mm}
    \caption{Classification results across all models and classes. We present F1-scores (mean $\pm$ std),$^{10}$ best-performing performance per class in bold, and the runner-up underlined. The second column indicates the size of the model counted in total parameters. The third column (FT) indicates task-related fine-tuning with the \textsc{Trump-2016} dataset.}
    \label{tab:main_results}
    \vspace{-3mm}
\end{table*}

\section{Experiments}
\label{sec:benchmarking}

\subsection{Methods}

\paragraph{Baseline} We consider two `naive' baselines: (a) a class-distribution random classifier (Dist.~Random), which samples predictions stochastically based on the class distribution of the training set, and (b) a linear SVM~\cite{cortes1995support} classifier using TF-IDF scores as features for 10K $n$-grams, where $n\in[1,2,3]$.\footnote{We present additional details in Appendix~\ref{sec:exp_details}.} The first one (Dist.~Random) sets a low bar considering the extreme data skewness; the second (TFIDF-SVM) helps us identify the degree to which the task can be solved mostly by identifying keywords and catch-phrases.

\paragraph{PLMs} We evaluate Pre-trained Language Models (PLMs), also known as BERT-like \cite{devlin-etal-2019-bert} encoder-only models. We use the large (approx.~350-400M parameters) versions of RoBERTa \cite{roberta-2019} and DeBERTa-V3 \cite{he2023debertav}. RoBERTa has been initially pre-trained with the Masked Language Modeling (MLM) objective, while DeBERTa-V3 uses an ELECTRA-like, Replaced Token Detection (RTD) objective, on large web corpora. We fine-tune both PLMs as 3-way multi-label classifiers using a shallow MLP on top of the final sentence representation.\footnotemark[7]

\paragraph{Open-weight LLMs} We also consider new-era LLMs. We use RLHF'd instruction-tuned versions of LLMs that exhibit strong performance on diverse NLP tasks in a zero-shot setting.
We benchmark a variety of publicly released open-weight models, specifically:
Meta's Llama 3.1 (8/70B) \cite{llama3}, Google's Gemma 3 (12/27B) \cite{gemma_2025}, and Alibaba's Qwen 3 (8/14/32B) \cite{qwen3technicalreport}.\footnote{We use Qwen 3 models in non-thinking mode.} We evaluate all models in a zero-shot setting and we also present results fine-tuning Llama 3.1 (8B) and Qwen 3 (8/14B) using a LoRA adapter~\cite{hu2022lora,dettmers2023qlora}. The base prompt we use is available in Appendix - Table \ref{tab:prompts}, dubbed as \textit{Base}.\footnotemark[7]

\paragraph{Proprietary LLMs} Finally, we consider two recent top-tier closed-source proprietary LLMs. We use OpenAI's GPT-4.1 -version~04/2025- \cite{gpt41} and Google's Gemini 2.5 Flash -version~05/2025- \cite{gemini} as a service (inference via their API).

\subsection{Main Results}
\label{sec:results}

Table~\ref{tab:main_results} presents the main results of our experiments, using F1-score on the test subset of Donald Trump's speeches (\textsc{Trump-2016}). A basic observation is that the classification task proves highly challenging across all examined methods. This finding aligns with previous work on coding populist discourse~\cite{bonikowski-2022}, which underscores the complexity of reliably identifying populist frames in political texts. The best performance is achieved by the fine-tuned RoBERTa classifier, demonstrating a macro-F1 score of 0.746. 

The fine-tuned RoBERTa and DeBERTa classifiers perform almost on par, substantially outperforming LLMs in a zero-shot setting (approx.~+20-30\%). We present additional details and experiments in different settings, i.e., data augmentation, and models with context, in Appendices~\ref{sec:exp_details} -\ref{sec:add_experiments}.

Open-weight LLMs struggle to perform in zero-shot settings, which include a brief concept definition and task description in the prompt.\footnote{See Table~\ref{tab:prompts} in the Appendix for details on prompts}\footnote{For each open-weight zero-shot model, we report the standard deviation ($\pm$ std) over multiple runs to account for prompt instability. We present additional details in Appendix~\ref{sec:exp_details}.}  The best-performing open-weight model, Qwen (14B), achieves a macro-F1 of 0.582. Overall, we observe that the Qwen 3 models outperform and are also more robust (low $\pm$ std), compared to Llama 3.1 and Gemma 3 models.

Similarly, top-notch proprietary LLMs demonstrate poor performance. The best-performing model, Gemini 2.5 Flash, achieves a macro-F1 of 0.574, scoring just below Qwen 3 (14B), which is the best-performing open-weight LLM. 

To render the comparison more comprehensive, we also fine-tune Llama 3.1 and Qwen 3 (8B, 14B) using LoRA adapters. Fine-tuning leads to substantial performance improvement, by approximately~20-25\%, although still substantially below the fine-tuned PLMs, except for Qwen 3 (14B). 

\section{Analyses}
\label{sec:llm_analysis}
Given the poor zero-shot performance of LLMs, we experiment with more elaborate prompting strategies, i.e., \textit{prompt tuning}, for the best-performing open-weight and proprietary LLMs (Section~\ref{sec:prompt_tuning}). We also present a \textit{meta analysis} where we reevaluate models based on a re-annotation of a part of the dataset by our domain expert (Section~\ref{sec:meta_evaluation}). We also examine \textit{word-specific relevance} scores to detect which (type of) words drive the models or lead to common errors (Section~\ref{sec:relevance}).

\subsection{Prompt Tuning}
\label{sec:prompt_tuning}
Our prompt tuning strategy includes augmenting the prompt and exploring the following settings, as summarized in Table~\ref{tab:prompts}: (a) \textbf{Baseline}, where we provide the model with the working definition of populism (Section~\ref{sec:intro}) and a brief task description (original prompt, as used to build Table \ref{tab:main_results}), 
(b) \textbf{Context-Aware}, where we present up to five preceding sentences from the same speech after the \emph{Baseline} prompt to augment the context, 
(c) \textbf{Distribution-Aware}, where we include the label distribution after the \emph{Baseline}~prompt (Table~\ref{tab:trump_dist}), (d) \textbf{K-shot}, where we explore few-shot prompting, and present K randomly-selected training examples, K/4 per category (incl.~examples with both people-centrism and anti-elitism), appended after the \emph{Baseline} prompt, (e) \textbf{RAG-Shot}, where the K demonstrated training examples are retrieved based on similarity to the target (examined) sentence.

\begin{table}[t]
    \centering
    \resizebox{\columnwidth}{!}{
    \begin{tabular}{ll|c|c|c|c}
         \multicolumn{2}{c|}{\textbf{Prompt Setting}} & \textbf{N} & \textbf{AE} & \textbf{PC} & \textbf{Avg} \\
         \midrule
         \multicolumn{2}{l|}{Baseline} & .916 & .491 & .340 & \cellcolor{c58} .582 \\
          \midrule
         \multicolumn{2}{l|}{Context-Aware} & .827 & .336 & .252 & \cellcolor{c47} .472 \\
         \multicolumn{2}{l|}{Distribution-Aware} & \underline{.932} & \underline{.517} & \underline{.358} & \cellcolor{c60} \underline{.602} \\
         \midrule
         \multirow{4}{4.5em}{K-Shot} & K=8 & \textbf{.943} & \textbf{.554} & \textbf{.386} & \cellcolor{c63} \textbf{.628} \\
         & K=16 & .932 & .520 & .392 & \cellcolor{c62} .615 \\ 
         & K=32 & .920 & .456 & .374 & \cellcolor{c58} .583 \\
         & K=64 & .929 & .487 & .375 & \cellcolor{c60} .597 \\
         \midrule
         \multirow{4}{4.5em}{RAG-Shot} & K=8 & \textbf{.943} & \underline{.528} & \underline{.377} & \cellcolor{c62} \underline{.616} \\
         & K=16 & .938 & .494 & .396 & \cellcolor{c61} .609\\
         & K=32 & .929 & .455 & .364 & \cellcolor{c58} .582 \\
         & K=64 & .926 & .438 & .356 & \cellcolor{c57} .573 \\
    \end{tabular}
    }
     \vspace{-1mm}
    \caption{Results of Qwen 3 (14B) performance across different prompt settings. \textit{Baseline} refers to the same prompt as in Table \ref{tab:main_results}.}
    \vspace{-1mm}
    \label{tab:llm_modes}
\end{table}

\paragraph{Qwen 3 -- Alternative Prompting}
In Table~\ref{tab:llm_modes}, we present prompt-tuning results for Qwen 3 (14B).\footnote{See results for Llama 3.1 (8B) in Appendix~\ref{sec:add_experiments_llama}, leading to similar observations.} 
Contrary to expectation, we observe that including preceding sentences as context (\emph{Context-Aware}) harms performance. We hypothesize that the model becomes heavily biased by the preceding sentences (e.g., when these are populist, and the examined is not) instead of using them as assistive context to disambiguate (e.g., to resolve co-references, as suggested by the prompt), leading to several misclassifications.
In contrast, there is a considerable performance improvement (approx.~+2\%) when we provide the expected label distribution (\emph{Distribution-Aware}), likely because the model is biased towards the majority category (neutral). 

Finally, few-shot prompting, where we provide a small number of training examples (\emph{K-Shot}, $K\!\!\in\!\![8,16]$), leads to a performance improvement of approx.~+3-4\%, while performance drops for larger $K\!\in\![32,64]$. Results from \emph{RAG-Shot} show a similar trend. Potentially, the model is unable to effectively perform long-context reasoning and utilize a larger set of demonstrations. Similarly, as $K$ increases, the input may include demonstrations that are highly contextual (populist or not in context), and thus unidirectionally bias the model.

\begin{table}[t]
    \centering
    \resizebox{\columnwidth}{!}{
    \begin{tabular}{lc|c|c|c|c}
         \multicolumn{2}{c|}{\textbf{Prompt Setting}} & \multicolumn{4}{c}{\textbf{Classes}} \\
         \textbf{K} & \textbf{Mode} & \textbf{N} & \textbf{AE} & \textbf{PC} & \textbf{Avg} \\
         \midrule
          \multirow{3}{1em}{0}  & Base & .903 & .513 & .306 & \cellcolor{c57} .574 \\
          & Think & .868 & .429 & .272 & \cellcolor{c52} .523 \\
          & Dist  & \underline{.931} & \underline{.557} & \underline{.372} & \cellcolor{c62} \underline{.620} \\
         \midrule
          \multirow{3}{1em}{32}  & Base & .912 & .558 & .329 & \cellcolor{c60} .600 \\
          & Think & .874 & .416 & .309 & \cellcolor{c53} .533  \\
         & Dist & \underline{.921} & \underline{.612} & \underline{.337} & \cellcolor{c62} \underline{.623} \\
         \midrule
         \multirow{3}{1em}{128} & Base & .925 & .597 & .351 & \cellcolor{c62} .624 \\
         & Think & .860 & .448 & .261 & \cellcolor{c52} .523 \\ 
         & Dist & \textbf{.928} & \textbf{.609} & \textbf{.346} & \cellcolor{c63} \textbf{.628} \\
    \end{tabular}
    }
    \vspace{-1mm}
    \caption{Results of  Gemini 2.5 Flash performance across different prompt settings. K refers to the number of presented examples for \emph{K-shot}.}
    \label{tab:gemini_modes}
    \vspace{-1mm}
\end{table}

\paragraph{Gemini 2.5 -- Prompt-tuning results}
Given the previous results for Qwen 3 (14B), we focus our analysis on \emph{Baseline} (here \emph{Base}), and further enhancement via \emph{K-Shot} (for $K\in[0,32,128]$) and \emph{Distribution-Aware} prompting for Gemini 2.5 Flash, the best-performing proprietary LLM. Furthermore, we evaluate Gemini's \emph{thinking} mode by instructing the model to generate up to 1000 ``thinking'' (reasoning) tokens before classifying the target sentence each time.\footnote{We reproduce the term as presented by~\citet{gemini}. The term refers to a prominent technique mostly known as ``chain-of-thought''~\cite{wei-et-al-2022} and later ``reasoning'' \cite{deepseekai2025}, where the model generates text that resembles its reasoning process before answering the task (question) at hand.} 

The results are presented in Table~\ref{tab:gemini_modes}. Comparing the performance of the base prompt across different Ks (0,32,128), we observe that \emph{K-shot} overall improves results by 3-5\%, while providing the label distribution alone leads to similar improvements. This aligns with our findings for Qwen 3 (14B). Nonetheless, contrary to Qwen 3 (14B), Gemini benefits from an increased number of demonstrations ($K\!=\!128$ vs 32 in \emph{Base} mode).
 
Surprisingly, enabling ``thinking'' (reasoning) leads to a clear performance drop across the board. We hypothesize that the model shifts to ``\emph{overthinking}'',  nudging it to label an additional 9\% of total sentences as populist (i.e., anti-elitist and/or people-centric) compared to the non-thinking mode, while only 3\% of sentences flip in the opposite direction. In short, ``thinking'' leads to over-identification of populist content.

Conversely, revealing the label distribution leads to the opposite phenomenon, inducing what we coin as ``\emph{hypervigilance}'', biasing the model toward the neutral (majority) class. In other words, when exposed to the label distribution, the model becomes noticeably more cautious when classifying sentences. In this setting, the distribution-aware model classifies an additional 5\% of sentences as neutral, with near zero flips in the opposite direction. This results in fewer misclassifications and therefore improves performance.

\begin{table}[t]
    \centering
    \resizebox{\columnwidth}{!}{
    \begin{tabular}{l|c|c|c|c|c||c}
         \textbf{Method} & \textbf{FT} & \textbf{N} & \textbf{AE} & \textbf{PC} & \textbf{Avg} & \textbf{Prev.} \\
         \midrule
         Students & - & .975 & .658 & .577 & \cellcolor{c74} .737 & -\\
         \midrule
         RoBERTa & \ding{52} & \textbf{.983} & .724 & \textbf{.695} & \cellcolor{c80} \textbf{.800} & \cellcolor{c75} \textbf{.746}\\
         Qwen 3 (8B) & \ding{52} & .972 & .653 & .565 & \cellcolor{c73} .730 & \cellcolor{c71} .714 \\
         \multirow{2}{*}{Qwen 3 (14B)} & \ding{56} & .951 & .612 & .374 & \cellcolor{c65} .646 & \cellcolor{c63} .628 \\
         & \ding{52} & .977 & \textbf{.730} & .598 & \cellcolor{c77} .768 & \cellcolor{c74} .740\\
         Gemini 2.5 & \ding{56} & .931 & .677 & .290 & \cellcolor{c63} .633 & \cellcolor{c63} .628 \\
    \end{tabular}
    }
    \caption{Results of the meta evaluation. We consider the best-performing settings for zero-shot models: Qwen 3 (14B \ding{56},\emph{K-Shot}, K=8), and Gemini 2.5 (\emph{K-Shot}, K=128 with distribution). The last column (Prev.) shows the original results comparing to students' annotations.}
    \label{tab:meta_analysis}
    \vspace{-3mm}
\end{table}

 \subsection{Meta Evaluation}
 \label{sec:meta_evaluation}

Determining whether a sentence is populist (or not) is no trivial task for models and humans alike.\footnote{See our remarks on \emph{Subjectivity} in Section~\ref{sec:limitations}.} First of all, a certain level of disagreement among human annotators is to be expected, as the task is based on \emph{interpreting} an abstract social science concept. Hence, to put the models' performance into perspective, we asked the domain expert of this project, who also trained the student annotators, to re-annotate the test set of \textsc{Trump-2016}, serving as an \emph{Oracle}. In this meta-evaluation, we re-assess the best-performing models (in their optimal configuration) against the expert-annotated test set (see Table \ref{tab:meta_analysis} for results). 

We treat the re-annotation as the gold standard and evaluate student annotations accordingly. The students achieve a macro-F1 of 0.737. However, they are outperformed by the RoBERTa model (macro-F1: 0.8), and the fine-tuned Qwen 3 (14B) (macro-F1: 0.768), and perform on par with the fine-tuned Qwen 3 (8B). Notably, the performance of the fine-tuned RoBERTa is improved by approx.~0.05, while the remaining models perform comparably to before. 
From a practical perspective, these results suggest that fine-tuned models, especially the RoBERTa classifier, achieve greater consistency than student annotators, effectively overcoming the "noise" introduced by non-expert labeling, thus exhibiting higher agreement with the domain expert. 

\subsection{Word-specific Relevance}
 \label{sec:relevance}
To better understand which words are most relevant for detecting populism, we take a closer look at the feature weights (coefficients) of the TFIDF-SVM model and relevance scores for the fine-tuned RoBERTa model per class.

We observe that the five most indicative expressions (n-grams) used by the TFIDF-SVM model for classifying a sentence as \textit{anti-elitist} are `donors', `establishment', `insiders', `rigged', and `special interests'. Respectively, the five most indicative expressions for \textit{people-centrism} are `the American people', `for you', `the people', and `forgotten'.

Next, we apply Layer-wise Relevance Propagation (LRP) \cite{ali2022xai} to the fine-tuned RoBERTa model. LRP is a feature attribution method that assigns a continuous (positive or negative) relevance score to each token in the input, reflecting its contribution to the predicted class label. From each correctly classified sentence, we extract the 5 "most relevant" tokens that were assigned positive scores and group them by part-of-speech tag (see Appendix Table \ref{tab:xai}). 

The results are consistent with the TFIDF-SVM coefficients: keywords such as `establishment', `rigged', and `special' are indicative of \textit{anti-elitism}, as well as `corrupt', `Washington', `Hillary', and `system'. There is also an overlap between the most relevant words for \textit{people-centrism} and \textit{anti-elitism}, such as `American'/`people', but also `government', `rigged', `corrupt', and `system'. Tokens most relevant to neutral sentences tend to be more generic, which is expected, as the model classifies these in the absence of explicit populist cues. 

When we conduct the same analysis on misclassified sentences, we find that similar keywords, such as `American', `people', `dishonest', `crooked', and `Hillary', frequently mislead the model into assigning populist labels.
Moreover, a manual inspection of the misclassified sentences reveals that RoBERTa tends to misclassify sentences as \textit{anti-elitist} when they include references to political corruption or the media, which are prominent in Donald Trump's speeches. Similarly, it tends to misclassify sentences with references to the 'American people' as \textit{people-centrism}. 

In both cases, the target sentences are indeed ambiguous and would require context from adjacent sentences to disambiguate. Lacking such context, the model tends to rely on lexical cues alone, particularly high-frequency keywords learned from populist training examples.

\section{Analysis of Donald Trump's Populism}
\label{sec:trump_analysis}

To demonstrate the potential of fine-tuned models in political science research, we apply our best-performing model, RoBERTa, to the extended dataset \textsc{Trump-Chronos}, which spans the years 2015-2025. Section~\ref{sec:pdi} introduces our basic metric, the Populism Discourse Index (PDI), while Section~\ref{sec:trump_hypotheses} presents our research questions and the results of our statistical analysis.

\subsection{Populism Discourse Index (PDI)}
\label{sec:pdi}

Although the sentence is our primary unit of analysis for NLP purposes, we propose a scoring scheme that assigns a discrete value to each complete speech ($S$), capturing the overall intensity of its populist content. Following prior work \cite{aslanidis2018measuring}, we refer to this per-speech metric as the \textit{Populism Discourse Index} ($\mathrm{PDI}$). As shown in the sections below, collecting $\mathrm{PDI}$ scores enables the statistical treatment required to answer the research questions of our case study.

We define the sentence-level scoring function as follows:\vspace{-1mm}
\begin{align}
\mathrm{Score}(s) =
  \begin{cases}
    0       & \quad \text{if } p(s) \text{ is N}\\
    1  & \quad \text{if } p(s) \text{ is AE $\lor$ PC} \\
    3   & \quad \text{if } p(s) \text{ is AE $\land$ PC} 
  \end{cases}
\label{for:pdi_sentence}
\end{align}
\noindent where $p$ is the model's prediction for a given sentence ($s$), N denotes a \textit{neutral} sentence, AE stands for \textit{anti-elitism}, and PC for \textit{people-centrism}. The \textit{exponential boost} assigned to \textit{fully populist} sentences (a constant set to 3 in our implementation) reflects the amplified cognitive and affective impact of archetypal populist declarations such as \emph{"The \underline{political elite} is the enemy of the \underline{American people}"} or \emph{"\underline{The system} must work for \underline{all Americans}, not just \underline{those at the top}"}. 

We further apply an \textit{adjacency multiplier} to pairs of consecutive sentences ($s_k$, $s_{k+1}$) that are labeled \textit{anti-elitist} (AE) and \textit{people-centric} (PC), or vice versa. This multiplier (set to 1.5) captures the rhetorical effect of back-to-back, yet distinct, populist claims. For instance, consider the following pair of adjacent sentences:\vspace{1mm}

\begin{tabular}{ll|l}
         \multicolumn{2}{c|}{$s$} & $p(s)$ \\
         \midrule
         $s_k$ &   \emph{"The system is rigged."} & AE \\
         $s_{k+1}$ & \emph{"The people must rise up."}& PC \\
\end{tabular}\\

\vspace{1mm}
Under simple additive scoring, the overall score for this sentence pair ($s_k$, $s_{k+1}$) would have been $1 + 1 = 2$ points,  but with the \textit{adjacency multiplier} the score is adjusted to $(1\times1.5) + (1\times1.5) = 3$ points. The intuition is that such sentence pairs effectively approximate the rhetorical force of a single, \textit{fully populist} sentence (AE $\land$ PC) from the audience's perspective.

In computing $\mathrm{PDI}$, we exclude sentences with fewer than three words, as these are typically interjections, e.g., \textit{"Wow!"}, \textit{"Incredible!"}, rhetorical prompts, e.g., \emph{"Right?"}, \emph{"Guess what!"} or are otherwise trivial. For similar reasons, we filter out sentences that begin with the string \emph{"Thank\underline{ }"}.\footnote{These filters do not bias our analysis; for reference, in our annotated dataset (\textsc{Trump-2016}), only 658 out of $\sim$15K sentences ($\sim$4\%) are part of the filtered out categories ($<3$ words or begin with ``Thank\underline{ }''), while only 3 of those 658 ($\sim$0\%) are labeled as populist (anti-elitist and/or people-centric).} After this preprocessing, the resulting dataset comprises 604,391 sentences, with $7.9\%$ of the original dataset excluded. Using Equation~\ref{for:pdi_sentence}, we define the speech-wise score ($\mathrm{PDI}_S$) as:
\begin{align}
    \mathrm{PDI}_S = \frac{1}{N} \sum_{s=i}^{N}{\mathrm{AdjScore}(s_i)} 
\label{eq:pdi}
\end{align}
\noindent where $\mathrm{AdjScore}$ is the adjusted scoring function, w.r.t. adjacent populist sentences, and $N$ is the total number of sentences ($s_i$) in a given speech ($S$).

Overall, our coding scheme leverages sentence-level granularity to compute a $\mathrm{PDI}_S$ for each of the 713 speeches in our dataset. Each speech is also tagged with metadata such as \textit{date} and \textit{location}, enabling its use as the unit of analysis in downstream statistical modeling. 

Compared to prior approaches that score populist content at the paragraph or speech level, our sentence-level framework avoids arbitrary segmentation and provides a more fine-grained and dynamic measure of populist discourse. 

\subsection{Research Questions \& Analysis}
\label{sec:trump_hypotheses}
To analyze Donald Trump's populist discourse, we address three central research questions (RQ):\vspace{2mm}

\noindent\rule{\columnwidth}{2pt}

\noindent\textbf{RQ1:} \emph{"Does Donald Trump's populism substantially fluctuate over time?"}

\noindent\rule{\columnwidth}{2pt}\\

\noindent\textbf{$\mathrm{H}_0$}: The intensity of Donald Trump's populist discourse does not change systematically across successive electoral campaigns.\\

\noindent\textbf{$\mathrm{H}_1$}: Donald Trump modulates the intensity of his populist discourse across electoral campaigns to serve strategic objectives.

\paragraph{Theoretical Rationale} The dominant ideational approach in populism studies defines populism as a ``thin-centered ideology'' (see `ideational approaches' in Section~\ref{sec:intro}). Under this view, populism constitutes the core worldview and authentic political identity of populist politicians. Accordingly, the intensity of their populist discourse should remain relatively stable across contexts ($\mathrm{H}_0$), reflecting ideological consistency. Rejection of $\mathrm{H}_0$ would suggest \textit{significant} temporal variation in populist discourse,  implying that strategic calculation, rather than ideological coherence, is the primary driver. Politicians guided by strategy should tailor populist appeals to audience receptivity, electoral competitiveness, or perceived efficacy in varying contexts ($\mathrm{H}_1$), in our case across campaigns (Table~\ref{tab:descriptive_stats}).

\paragraph{Operationalization \& Methodology} Under $\mathrm{H}_0$, we expect no statistically significant differences in mean $\mathrm{PDI}$ across campaign periods, with any variation attributable to sampling error. Under the alternative, $\mathrm{H}_1$, we expect systematic differences, reflecting temporal variation in populist intensity. 

To test $\mathrm{H}_0$, we employ a hierarchical parametric testing framework, using the individual speech as the unit of analysis and campaign period as the primary grouping variable. The dependent variable is the $\mathrm{PDI}$ score (Equation~\ref{eq:pdi} - Section~\ref{sec:pdi}) calculated for each speech delivered during three U.S. presidential campaigns (2016, 2020, 2024). For comparative purposes, we include Trump's speeches during the Republican primaries (2015-2016). Each campaign period is delineated from the official candidacy declaration to election day; the 2016 primaries end with Trump’s nomination at the Republican National Convention. Descriptive statistics are presented in Table~\ref{tab:descriptive_stats}. We conduct a one-way analysis of variance (ANOVA), followed by pairwise comparisons using independent-samples t-tests across all campaign periods.

\begin{table}[t]
    \centering
    \resizebox{\columnwidth}{!}{
    \begin{tabular}{c|r|c}
         \textbf{Campaign Name} & \textbf{\#N} & $\bm{\overline{\mathrm{PDI}}(\pm)}$ \\
         \midrule
         \textbf{2016 Primaries} & \multirow{2}{*}{220}  & \multirow{2}{*}{1.67 $\pm$ 2.37}\\
         (16JUN2015--19JUL2016) & \\
         \textbf{2016 Election} & \multirow{2}{*}{133} & \multirow{2}{*}{6.30  $\pm$ 4.92} \\
         (21JUL2016--08NOV2016) & \\
         \textbf{2020 Election} & \multirow{2}{*}{98} & \multirow{2}{*}{2.44  $\pm$ 2.01} \\
         (18JUN2019--03NOV2020) & \\
         \textbf{2024 Election} & \multirow{2}{*}{141} & \multirow{2}{*}{1.50  $\pm$  1.03} \\
         (15NOV2022--05NOV2024) & \\
    \end{tabular}
    }
    \vspace{-3mm}
    \caption{Descriptive statistics across Trump’s campaign periods. \#N refers to the number of speeches, and $\overline{\mathrm{PDI}}$ to the mean $\mathrm{PDI}$ across all speeches.}
    \label{tab:descriptive_stats}
    \vspace{-4mm}
\end{table}

\paragraph{Results}
The one-way ANOVA results decisively reject the null hypothesis ($\mathrm{H}_0$), indicating a significant effect of campaign period on $\mathrm{PDI}$ scores, \textit{F}(3, 588) = 85.141, $p\!<\!0.001$, with a large effect size ($\eta^2\!=\!0.303$). The data reveal a clear pattern: Trump's populist discourse peaks dramatically during the 2016 general election ($\overline{\mathrm{PDI}}\!=\!6.30$) compared to the 2016 primaries ($\overline{\mathrm{PDI}}\!=\!1.67$), representing a 277\% increase that is highly statistically significant ($t\!=\!-11.867$, $p\!<0.001$). Subsequent campaigns show statistically significant declines: 2020 $\overline{\mathrm{PDI}}\!=\!2.44$ and 2024 $\overline{\mathrm{PDI}}\!=\!1.50$, both with ($p\!<\!0.001$). Pairwise comparisons (See Table~\ref{tab:pairwise_ttests_pdi} - Appendix~\ref{sec:trump_analysis_add}) confirm that these differences are not only statistically significant but also substantively meaningful, with large effect sizes. 

These systematic shifts strongly support $\mathrm{H}_1$: Trump strategically adjusts the intensity of his populist discourse across campaigns, with the 2016 general election marking a clear inflection point where populist rhetoric was maximally weaponized before being moderated in later cycles.
Notably, our bonus scoring system (1.5$\times$ multiplier for adjacent AE–PC sentence pairs) affects only 0.06\% of sentences, indicating that results are robust to adjacency effects. 

However, our findings may not be reflecting the real \textit{volume} of populist discourse in a given speech, due to the nature of our scoring scheme and the use of the sentence as the unit of analysis: short, punchy populist sentences (e.g. \textit{"The system is rigged!"} or \textit{"Elites betray us!"}) can inflate the $\mathrm{PDI}$ score, even if the rest of the speech is predominantly composed of long, neutral sentences. 

To address this concern, we switch from measuring $\mathrm{PDI}$ solely on sentence frequency (Equation~\ref{eq:pdi} - Section~\ref{sec:pdi}) to accounting for content volume by introducing a refined scoring scheme, dubbed $\mathrm{WPDI}$ (Weighted Populist Discourse Index). $\mathrm{WPDI}$ adjusts $\mathrm{PDI}$ by the relative length of populist (AE and/or PC) versus neutral (N) sentences in each speech (Equation~\ref{eq:wpdi}). Specifically, $\mathrm{WPDI}$ normalizes $\mathrm{PDI}$ by the ratio of average populist sentence length ($\bar{L}_{AE \vee PC \vee (AE \wedge PC)}$) to average neutral sentence length ($\bar{L}_{N}$), using word count as length, i.e.:
\begin{align}
\text{WPDI}_S = \text{PDI}_S \times \frac{\bar{L}_{AE \vee PC \vee (AE \wedge PC)}}{\bar{L}_{N}}
\label{eq:wpdi}
\end{align}

\begin{table*}
\centering
\resizebox{0.9\textwidth}{!}{
\begin{tabular}{c|p{4cm}|c|c||p{4cm}|c|c}
\textbf{Campaign} & \multicolumn{2}{c|}{\textbf{Ballotpedia Swing States}} & \textbf{Speeches} & \multicolumn{2}{c|}{\textbf{High-Attention States}} & \textbf{Speeches} \\
\midrule
\textbf{2016} & AZ, CO, FL, IA, MI, NV, NH, NC, OH, PA, VA, WI & 12 & 123 vs 10& FL, NC, OH, PA, CO & 5  & 81 vs 52\\
\midrule
\textbf{2020} & AZ, FL, GA, IA, MI, MN, NV, NH, NC, OH, PA, TX, WI & 13 & 82 vs 16& PA, NC, FL, MI, WI, AZ, MN, OH & 8 & 68 vs 30\\
\midrule
\textbf{2024} & AZ, GA, MI, NV, NC, PA, WI & 7 & 73 vs 68& IA, PA, NC, NH, MI, WI & 6 & 89 vs 52\\
\end{tabular}
}
\caption{Swing State Clustering Scheme by Campaign: Speech counts shown as swing vs. non-swing states.}
\label{tab:swing_states}
\end{table*}

This adjustment ensures that the actual verbal footprint of populist content is reflected more accurately compared to relying on the frequency of populist sentences.

Running the same analysis with $\mathrm{WPDI}$ confirms and strengthens our findings. The ANOVA figure rises to $F\!=\!86.618\!>\!85.141$ and  pairwise comparisons yield larger effect sizes (Appendix~\ref{sec:trump_analysis_add} - Table~\ref{tab:pairwise_ttests_wpdi}). Correlation between $\mathrm{PDI}$ and $\mathrm{WPDI}$ is extremely high at both the campaign-period ($r\!=\!0.986$) and speech ($r \!=\!0.98$) level, suggesting that both metrics capture the same underlying dynamics, with $\mathrm{WPDI}$ offering greater sensitivity. However, we suggest using $\mathrm{PDI}$ in future work, due to its simpler and more intuitive nature.

\paragraph{Discussion} Our findings challenge binary conceptions of populism that dominate academic and public debates. Rather than a fixed ideological trait, populist discourse emerges here as a variable mode of political meaning-making, modulated across time and contexts. 

This supports the idea that scholars should shift from rigidly typological to dynamic, \textit{gradational} approaches in populism research. The substantial temporal variation in Donald Trump's populist discourse that we document across the 2016 primaries and the three presidential campaigns highlights the inadequacy of categorical approaches that rely on a static classification of  political phenomena as 'populist' or 'non-populist.' Our sentence-level, metrics-based framework allows for fine-grained tracking of populist discourse and captures strategic modulation often overlooked by binary coding schemes. This alternative methodological approach addresses the growing concern that the term 'populist' is overused and analytically imprecise. 

While our analysis focuses on a single politician, Donald Trump represents a paradigmatic case of modern populist leadership. If such a central figure exhibits strategic fluctuation in populist rhetoric, it suggests that similar dynamics may be at play in other, as yet unexplored, cases, underscoring the need for comparative research across political contexts to substantiate this claim.

\noindent\rule{\columnwidth}{2pt}

\noindent\textbf{RQ2:}  \emph{"Does Trump's populist discourse intensity vary between swing states and non-swing states?"}

\noindent\rule{\columnwidth}{2pt}\\

\noindent Building on the rejection of $\mathrm{H}_0$ in RQ1, which points to the strategic use of populist discourse, we further explore whether such a strategy includes geographic targeting. Specifically, we examine speeches delivered by Donald Trump in key battleground states during the 2016, 2020, and 2024 presidential campaigns.\\

\noindent\textbf{$\mathrm{H}_2$}: There is no difference in populist discourse intensity between swing and non-swing states.\\

\noindent\textbf{$\mathrm{H}_3$}: Trump intensifies populist discourse in swing states to maximize electoral returns.

\paragraph{Theoretical Rationale} Rational political actors are expected to allocate finite campaign resources where the expected return is highest \cite{bartels1985resource}. In U.S. presidential elections, due to the winner-take-all allocation of electoral votes at the state level, candidates focus disproportionately on key battlegrounds, or \emph{swing states}, to secure Electoral College majorities \cite{shaw2024battleground}. Given the theorized effectiveness of populist discourse in mobilizing voters through emotional appeals and in-group/out-group distinctions \cite{mudde2017populism,Norris_Inglehart_2019, dai2022politicians}, we hypothesize that Trump deploys more intense populist rhetoric in swing states, where voter mobilization is most consequential. We therefore test hypothesis $\mathrm{H}_2$ against the alternative, $\mathrm{H}_3$.

\paragraph{Operationalization \& Methodology} Due to the lack of consensus on swing state designations across election cycles, we test two different clustering (classification) schemes to define swing states: (a)  Ballotpedia's\footnote{\url{https://ballotpedia.org}} campaign-specific lists based on pre-election polling data, and (b) a data-driven proxy we term \textit{High Campaign Attention States}, defined as states in the top 25th percentile of Trump campaign speeches during each election cycle (Table~\ref{tab:swing_states}). The latter reflects revealed preferences in campaign resource allocation. For each scenario, we calculate both $\mathrm{PDI}$ and $\mathrm{WPDI}$ scores for all Trump speeches delivered in swing states ($S$) versus non-swing states ($\neg S$) during the three presidential campaigns.

\begin{figure}[t]
    \centering
    \resizebox{\columnwidth}{!}{
    \includegraphics{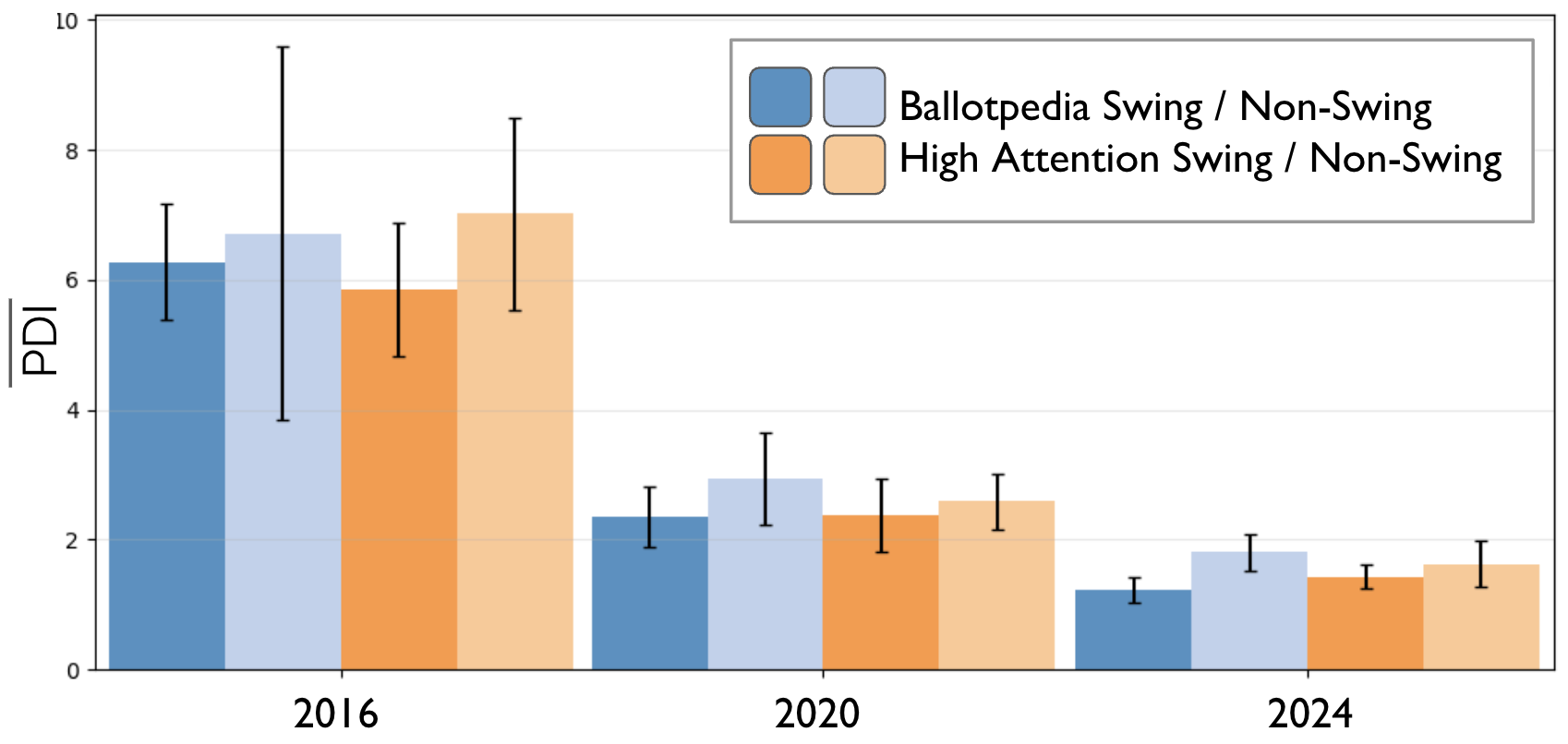}
    }
    \vspace{-4mm}
    \caption{$\overline{\mathrm{PDI}}$ scores by campaign and swing state type.}
    \label{fig:pdi_swign_states}
    \vspace{-4mm}
\end{figure}

We conduct independent-samples t-tests to compare mean $\mathrm{PDI}$ and $\mathrm{WPDI}$ scores between swing and non-swing states per campaign and clustering method. For each test, we calculate Cohen’s \textit{d} to assess effect size, and construct confidence intervals for group means ($\overline{\mathrm{PDI}}$), depicted in Figure~\ref{fig:pdi_swign_states}. The two clustering approaches serve as robustness check for our operational definition of swing states.

\paragraph{Results} Out of 12 independent-samples t-tests, only 2 (16.7\%) yielded statistically significant results, both in the 2024 campaign under the Ballotpedia definition (Appendix~\ref{sec:trump_analysis_add} - Table~\ref{tab:q2_ttest_results}), which means that overall we cannot reject $\mathrm{H}_2$. Interestingly, in the two statistically significant cases, Trump used \textit{less} populist rhetoric in swing states than in non-swing states. Specifically, for $\mathrm{PDI}$, $\overline{\mathrm{PDI}_S}\!=\!1.226$ vs $\overline{\mathrm{PDI}_{\neg S}}\!=\!1.804$ ($p\!<\!0.001$, $d\!=\!-0.579$); for $\mathrm{WPDI}$,  $\overline{\mathrm{WPDI}_S\!}=\!1.607$ vs $\overline{\mathrm{WPDI}_{\neg S}}\!=\!2.470$ ($p\!<\!0.001$, $d=\!-\!0.588$).\footnote{See Tables~\ref{tab:q2_descriptive_stats}-\ref{tab:q2_ttest_results} for detailed results. Also notice that mean $\mathrm{PDI}$/$\mathrm{WPDI}$ is lower in swing states compared to non swing states across all scenarios.} The remaining 10 comparisons showed no significant differences, with effect sizes ranging from negligible to small ($d\!=\!0.014-0.291$).\footnote{We do not correct for multiple comparisons, as each campaign constitutes a distinct hypothesis test. In any case, applying a Bonferroni correction within each campaign ($\alpha=0.05/4=0.0125$)  confirms the significance of the 2024 Ballotpedia results ($p<0.001$ for $\mathrm{PDI}$ and $\mathrm{WPDI}$).}

\paragraph{Discussion} Our findings suggest that Trump’s populist discourse was not geographically targeted in a systematic manner. The only observed significant differences run counter to expectations, with \textit{lower} levels of populist rhetoric in swing states during 2024. This may reflect a strategic shift toward moderation in competitive contexts, where aggressive populist messaging risks alienating the crucial pool of undecided voters \cite{dai2022politicians}. Future comparative research should investigate whether other populist figures show similar patterns or if strategic geographic variation in populist rhetoric is context-dependent.\\\vspace{-2mm}

\noindent\rule{\columnwidth}{2pt}

\noindent\textbf{RQ3:}  \emph{"How does Donald Trump's populism spread out within a speech?"}

\noindent\rule{\columnwidth}{2pt}\\

\noindent Our final research question examines the strategic placement of populist invocations within speeches.\\

\noindent\textbf{$\mathrm{H}_4$: }Populist discourse is uniformly distributed throughout a Trump speech.\\

\noindent\textbf{$\mathrm{H}_5$:} Donald Trump bookends his speeches with populist content to maximize efficacy.

\paragraph{Theoretical Rationale} Political speeches typically address a variety of themes, with much of the content devoted to policy issues and programmatic commitments. According to collective action frame theory as applied to populist discourse, populist actors perform three sequential framing tasks to achieve mobilization: (a) the \textit{diagnostic task}, which identifies the problem and attributes blame ("what is the problem?"; "who is responsible?"); (b) the \textit{prognostic task}, which outlines potential solutions ("what can be done?"); and (c) \textit{the motivational task} which calls supporters to action, urging them to join a movement or vote for a candidate. Qualitative research \cite{aslanidis-2024} suggests that populist discourse excels in diagnostic and motivational framing, but underperforms in the prognostic phase, due to its inability to offer solutions that satisfy the policy aspirations of its typically heterogeneous audience. Therefore, we expect Trump to cluster populist discourse at the beginning of his speeches for diagnostic framing, and at the end to deliver motivational appeals. Contrary, the middle segments, associated with prognostic framing, should feature notably less populist content. 
 
\paragraph{Operationalization \& Methodology} We analyze sentence-level positioning across 654,409 sentences drawn from 707 Trump campaign speeches. Unlike RQ1 and RQ2, we do not apply sentence-level filters; however, we exclude six speeches with $\mathrm{PDI}$ = 0. To test our hypotheses, we employ
a 3-bin scheme using a balanced 20-60-20 split (Opening 20\%, Body 60\%, Closing 20\%), with the body segment corresponding to where programmatic content is typically concentrated. We compute the \textit{Populist Volume} (PV) per bin, i.e. the proportion of populist sentences ($S^P_i$) in a given bin ($bin_i$) relative to all \textit{populist} sentences ($S^P$) in the speech:
\begin{equation*}
    \text{PV}(bin_i) = \frac{S^P_i}{S^P}
\end{equation*}

We use the $\mathrm{PV}$ metric to directly evaluate the bookending hypothesis ($\mathrm{H}_5$) by identifying where populist content is most concentrated. We assess three categories: \textit{overall populism} (AE$\vee$PC), \textit{anti-elitism} (AE), and \textit{people-centrism} (PC). Sentences that the model has tagged as both \textit{anti-elitist} and \textit{people-centric} (AE$\wedge$PC) contribute equally to each category's analysis. In order to assess the bookending hypothesis, we carry out pairwise paired t-tests between all three bins. While carrying out significance tests, we reweigh the ratio in the 3-bin scheme and also account for multiple comparisons with a Bonferroni correction of $\alpha$ levels.

\begin{figure}[t]
\centering
    \resizebox{0.95\columnwidth}{!}{
    \includegraphics{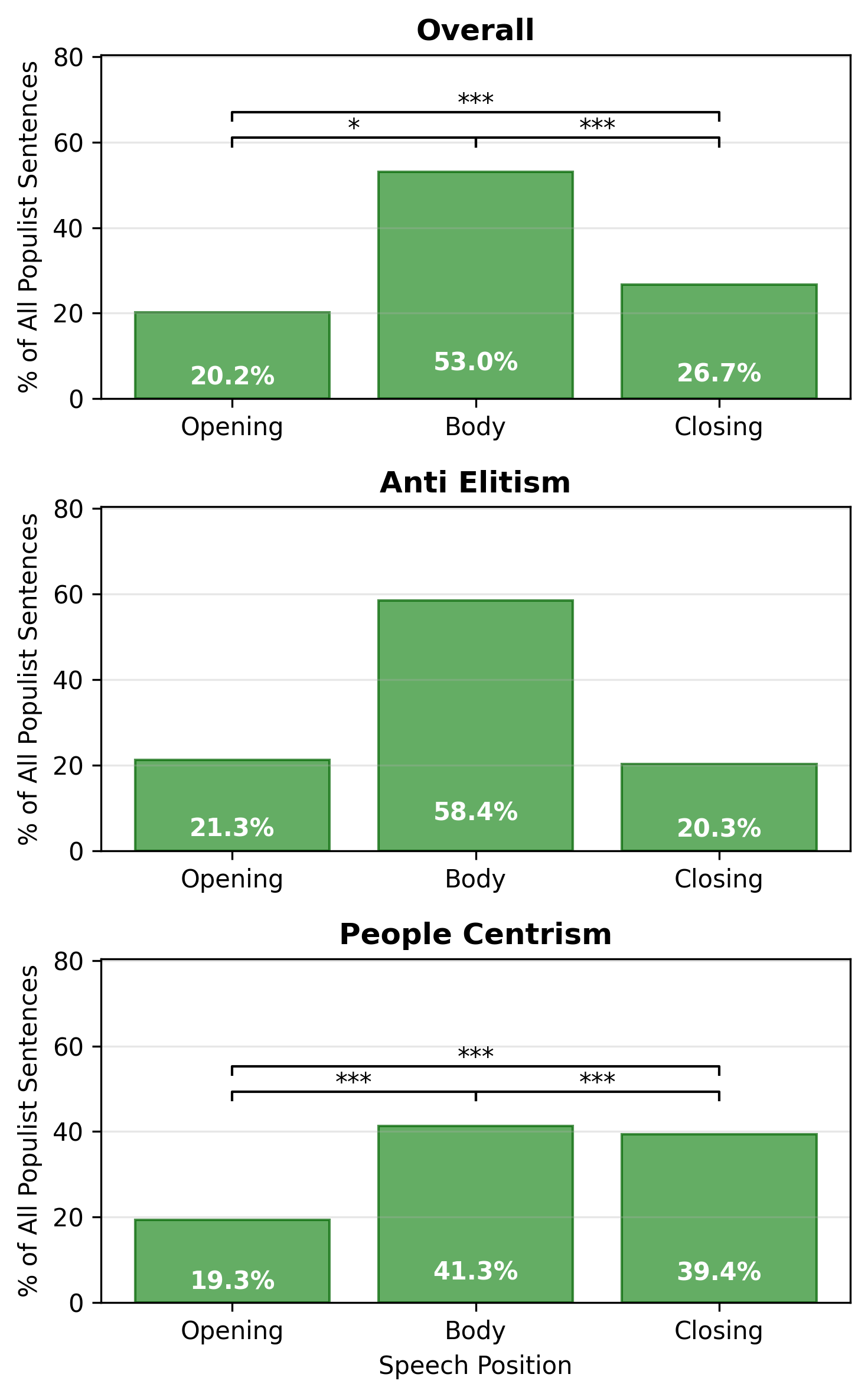}
    }
    \vspace{-5mm}
    \caption{Distribution of populist content across speech. Bars depict $\mathrm{PV}$ (\% among all populist sentences) for a 3-bin scheme using a balanced 20-60-20 split. Significance levels of comparisons between the opening/closing bin and all other bins are depicted as *** (p<0.001), * (p<0.05). No connections means that the comparison is not significant.}
    \vspace{-5mm}
    \label{fig:speech_5_binsenter-label}
\end{figure}

\paragraph{Results} We find significant differences for \textit{overall populism} and \textit{people-centrism} between all bins (Appendix~\ref{sec:trump_analysis_add} - Table \ref{tab:opening_closing_analysis}). As shown in Figure \ref{fig:speech_5_binsenter-label}, populist discourse is patterned, with approx.~$27\%$ of all populist sentences appearing in the closing of speeches compared to approx. $20\%$ in the opening. \textit{People-centrism} is heavily back-loaded, with nearly 40\% of people-centric sentences appearing in the closing of speeches compared to approx.~$19\%$ in the opening. We do not find significant differences for \textit{anti-elitism}.

Overall, the results give partial support to our alternative hypothesis {$\mathrm{H}_5$}. Anti-elitism does not contribute to the populist bookending of Trump speeches, however, people-centrism is prominent in the initial, diagnostic phase, and even more so in the closing, motivational segments of his speeches.

\paragraph{Discussion} Our findings reveal clear positional patterns in Trump's populist discourse, with strong back-loading of people-centrism. People-centric appeals emerge as a key feature of motivational framing, with usage rates doubling from opening to closing. 
The middle sections of speeches exhibit reduced populist content, offering partial support for the broader book-ending hypothesis.

This temporal separation suggests purposeful rhetorical sequencing, with people-centrism concentrated in the closing segments to mobilize and galvanize supporters. This pattern is consistent in over 700~speeches, indicating deliberate design rather than incidental variation. However, generalizability remains uncertain. For instance, people-centric appeals may naturally cluster toward closings in many types of political discourse, regardless of populist content. Future research should examine whether this pattern is unique to populist framing or reflects broader rhetorical conventions. 

\begin{figure}
    \centering
    \includegraphics[width=\columnwidth]{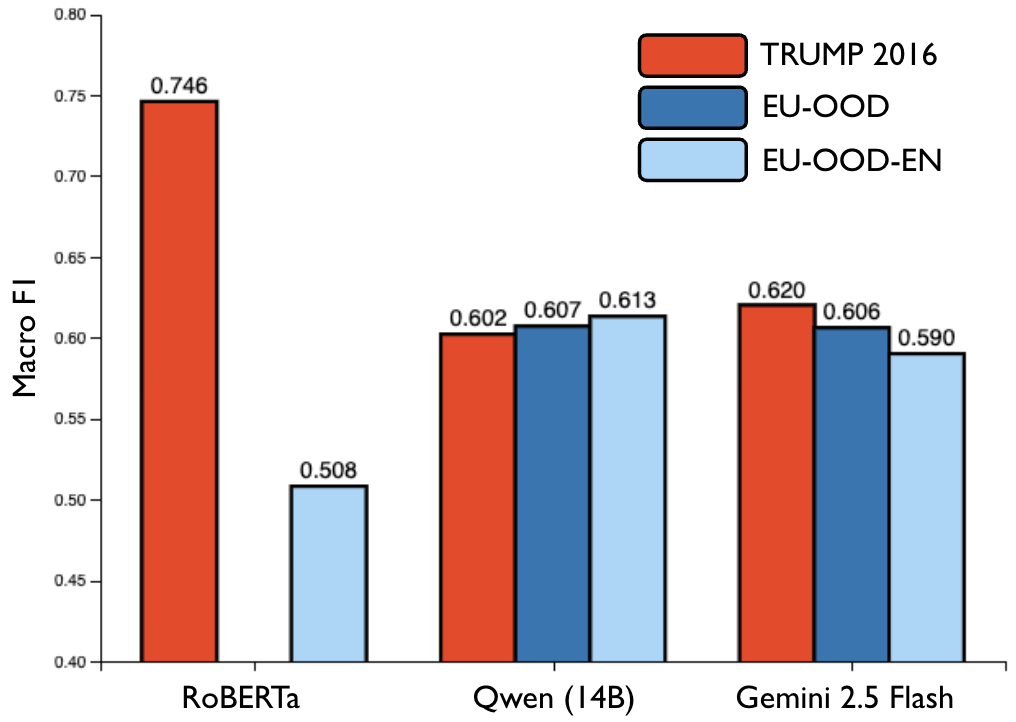}
    \vspace{-8mm}
    \caption{Performance of the best-performing examined models on the \textsc{EU-OOD} dataset in original languages and translated versions (-EN).}
    \vspace{-4mm}
    \label{tab:eu_speeches}
\end{figure}

\section{OOD Exploration -- The case of Europe}
\label{sec:eu_benchmarking}

Given that our annotated dataset (\textsc{Trump-2016}) is limited to a single individual and the specific historico-political context of the 2016 U.S. presidential election, we evaluate the models' performance in an Out-of-Domain (OOD) setting. To this end, we curate and annotate a small set of speeches by European politicians (\textsc{EU-OOD}), as described in Section~\ref{sec:datasets}. We assess the best-performing models: RoBERTa, Qwen 3-14B, and Gemini 2.5 Flash. For Qwen and Gemini, we run zero-shot experiments by adding the label distribution from the original dataset (\textsc{Trump-2016}) to the prompt (\textit{Distribution-Aware} from Tables \ref{tab:llm_modes}-\ref{tab:gemini_modes}), as a minimal intervention informing on data skewness. 

Figure~\ref{tab:eu_speeches} reports results using both the original-language sentences (\textsc{EU-OOD)} and their English translations (\textsc{-EN}). We observe that both LLMs (Qwen 3 and Gemini 2.5 Flash) outperform the fine-tuned RoBERTa classifier by approximately 10\%. In contrast, RoBERTa's performance drops by roughly~24\% in the OOD setting, while both LLMs show stable performance across datasets. This is a clear indication that our classifier has substantially overfitted Donald Trump's populist discourse. Furthermore, performance is comparable using original or translated (-EN) sentences.

Since our \textsc{EU-OOD} dataset is fairly small (six speeches by five politicians), we omit per-speaker analyses from the main text of this study. However, we point out that, in some speeches, the coding leads to very few sentences as populist (\textit{anti-elitist} and/or \textit{people-centric}), raising doubts on whether these politicians are properly designated as populists in the academic literature.\footnote{We present results in the Appendix~\ref{sec:per_eu_speaker} - Tables~\ref{tab:eu_speeches_extra}-\ref{tab:eu_speeches_percentage}.}

\section{Limitations \& Open Challenges}
\label{sec:limitations}
\paragraph{Annotated Trump Dataset} Our newly released annotated dataset, \textsc{Trump-2016} (Section~\ref{sec:datasets}), is relatively small (70 speeches, $\sim$15K sentences) and focuses exclusively on a single individual, Donald Trump, within the specific context of the 2016 U.S. presidential election. As a result, models trained on this dataset, such as RoBERTa, adopt a narrow understanding of populism and struggle to generalize across contexts (see Section~\ref{sec:eu_benchmarking}). Accordingly, our analysis (Section~\ref{sec:trump_analysis}) is confined to Trump, and our findings should not be generalized to other political figures. In future work, we plan to expand our datasets and analyses to include a more diverse set of political actors across different ideological orientations, geographic regions, and languages.

\paragraph{Augmenting Context} In our study, we primarily classify sentences in isolation, i.e. out-of-context, which limits our ability to incorporate relevant contextual cues, such as surrounding sentences. In many cases, populist discourse is context-dependent, and key interpretive clues are referentially underspecified, often taking the form of pronouns or deictic expressions. For example, in the sentence \emph{"\underline{They} are acting against \underline{you}!"}, the referents of `they' and `you' may carry strong political significance but cannot be identified (disambiguated) without additional context. 

As shown in Section~\ref{sec:prompt_tuning}, providing up to five preceding sentences as context to Qwen 3 (14B). and to Llama 3.1 (8B) in Appendix~\ref{sec:add_experiments_llama}, hurts classification performance. In preliminary experiments, we also considered a variant of the fine-tuned RoBERTa classifier that is enhanced with additional context. Specifically, we augmented the input with up to 7 preceding sentences separated by the special \texttt{<sep>} token (Appendix~\ref{sec:hier_roberta}), but this yielded no performance gain. Similarly, we tested Gemini 2.5 Flash with the full speech as context and observed negative results. 

These negative results are a strong indicator that models cannot effectively utilize contextual information and are rather negatively biased (misguided). Going forward, we plan to explore alternative forms of contextual augmentation or consider using substantially larger models that may leverage extended context effectively and faithfully.

\paragraph{Subjectivity} The concept of populism is heavily debated in both academia and the media (see Section~\ref{sec:intro}). Identifying and classifying populist discourse is inherently subjective and often context-sensitive, much like many other social science concepts. Therefore, annotating populist discourse is a labor-intensive process that demands substantial training resources to achieve acceptable inter-annotator agreement. In our experience, two issues posed particular challenges for human annotators: (a) negative references to the media, and (b) references to corruption. Both can qualify as anti-elitist, depending on framing and context. For instance, if the media are framed as part of an establishment pitted against the people, such references should be coded as anti-elitist. However, this framing is not always present. Similarly, anti-corruption discourse qualifies as populist, provided that the allegedly corrupt actors are discursively framed as members of the elite. Yet in many instances, corruption references lack such framing. These challenges again highlight the importance of contextual interpretation when identifying populist discourse. Overall, a comprehensive study should complement computational results with qualitative insights to draw robust conclusions about the populist character of a given political phenomenon.

\section{Related Work}

Below, we summarize a selection of works on populism detection employing NLP methods:

\citet{bonikowski-2022} investigate populist, nationalist, and authoritarian discourse in U.S. presidential campaign speeches (1952-2020). Populism is defined as ``\emph{a form of moral claims-making that juxtaposes a fundamentally corrupt elite with the virtuous people and promises to restore political power to the latter}''. They deploy a RoBERTa classifier, which they fine-tune in an active learning fashion, based on manually annotating 2,224 paragraphs in total across six non-mutually exclusive frames, including populism, which they code in binary terms, e.g., populist or not. Speeches are binarized by labeling them as populist if at least one paragraph contains populist appeals. Paragraphs were selected as the unit of analysis as sentences were deemed too granular, while full speeches were considered as having a low signal-to-noise ratio for effective classification. The fine-tuned RoBERTa classifiers substantially outperform Random-Forest classifiers, however, the authors do not compare the performance of their model against modern instruction-tuned LLMs, as their work predates the widespread availability of such models.

\citet{Erhard_2025} develop PopBERT, a transformer-based multi-label classifier for detecting populist discourse in German parliamentary speeches. Following the ideational definition of populism, they annotate 8,795 sentences from the German Bundestag (2013-2021) for anti-elitism and people-centrism, with moralizing language as a necessary condition for populist classification (according to the authors, this allows them to distinguish between rational critiques of elites and populist language \textit{per se}). Their approach extends beyond core populist dimensions to identify ``host ideologies'' (left-wing socialism vs. right-wing nativism) that attach to populist rhetoric, according to their working definition. They fine-tune a German BERT model using active learning for data sampling and employ a highly permissive annotation aggregation strategy where any sentence labeled populist by at least one of five coders is considered positive, leading to low inter-annotator agreement.

\citet{klamm-etal-2023-kind} develop MoPE (Mentions of the People and the Elite), a cross-lingual dataset for entity-level detection of populist references in parliamentary debates. Following the ``thin populism'' framework, they create a hierarchical annotation scheme to identify mentions of ``the People'' and ``the Elite'' in text, regardless of specific terminology used. They annotate 9,297 mentions across German Bundestag speeches (2017-2021) and 1,423 mentions in English EuroParl data, training transformer-based models to automatically detect these entity references. While complementary to our work, their approach focuses on identifying specific entity mentions (e.g., ``taxpayers'', ``government'') rather than classifying populist discourse patterns at the sentence level.

\citet{zhang_2024} detect populist discourse in Chinese social media by analyzing over 100K posts as their unit of analysis. They employ manual annotation by five coders for binary classification (populist vs. non-populist) to train a Chinese RoBERTa model. Their methodology adapts Mudde's definition of populism to the Chinese context as an intersection of populism and ultranationalism, targeting foreigners, Westerners, pro-Western intellectuals, and political elites. Using topic modeling and correspondence analysis, they identify distinct populist themes including ``corruption'', ``betrayal'', ``super-national treatment'', and ``immoral West''. Their work appears to primarily capture nationalist sentiment rather than core populist discourse, as evidenced by their focus on cultural invasion themes and foreign enemies.

\citet{Halterman_2025} introduces a methodology for using LLMs to generate synthetic training data to later train supervised text classifiers rather than employing LLMs directly for annotation tasks. In a populism-focused validation, among other use cases, the author uses GPT-3.5 to generate 5,357 synthetic populist manifesto statements across 27 European countries in 22 languages, drawing on Mudde's thin definition of populism to conceptualize populism as a conflict between ``good, common people'' and ``out-of-touch, self-serving elites''. The synthetic sentences, along with 36,509 negative examples generated from policy position descriptions, are used to train a sentence-level binary classifier using the SetFit few-shot framework without any gold-standard hand-labeled training data.

\citet{tao2025measuring} examine online populist discourse in China. Adhering to Laclau's discursive approach, they examine invocations of popular will against ``conspiring elites''. Their methodology relies on Semantic Role Labeling (SRL) techniques to extract semantic triplets from text and a rule-based method to classify triplets based on a curated dictionary. They then use this tool to examine the operationalization of people-centrism and anti-elitism, considering three trending events that occurred in China between 2019 and 2021.

\section{Conclusion}

Focusing on Donald Trump’s campaign speeches, a politician widely recognized as a populist, we show that a fine-tuned RoBERTa classifier outperforms both zero-shot and fine-tuned LLMs, including open-weights models like LLaMA 3.1, Qwen 3, and Gemma 3, as well as proprietary models such as GPT-4.1 and Gemini 2.5 Flash, in a multi-label, multi-class classification setup. 
We find that providing information about the label distribution, particularly highlighting the skew toward non-populist sentences, improves zero-shot performance. Providing relevant examples for each class also helps, while activating ‘reasoning’ mode in models such as Gemini 2.5 Flash appears to reduce performance. 

Deploying RoBERTa to a larger, unlabeled collection of Trump's campaign speeches from 2015 to 2025 reveals a decline in populist rhetoric over time, which we attribute to strategic calculation. However, we detect no meaningful difference in populist intensity between swing and non-swing states. On the other hand, we do uncover a clear temporal structure in Trump’s speeches where populist, and more specifically, people-centric appeals, are concentrated toward the end. 

Finally, we conduct a small-scale out-of-distribution (OOD) exploratory evaluation using European political speeches. While RoBERTa’s performance drops considerably in this setting, Qwen 3 and Gemini 2.5 achieve comparable results and outperform RoBERTa by approximately 10\%, suggesting superior robustness across domains.

In summary, while LLMs continue to struggle with the ambiguity and contextual demands of detecting populist discourse, the best-performing models achieve a level of performance comparable to that of human annotators. Future research should explore more sophisticated prompt tuning techniques and apply these models to larger, more diverse datasets with high-quality gold annotations.\vspace{2mm}

\section*{Author Contributions}

\begin{itemize}[itemsep=-0.2em,leftmargin=13pt]
    \item The first co-author (IC) was responsible for conducting most of the experiments in Sections 3 and 4, including fine-tuning PLMs and open-weight LLMs, designing the prompt-tuning experiments and running those for the open-weight LLMs; evaluating the performance of all models; curating the publication of resources; and co-writing the paper. 
    \item The second co-author (SB) was responsible for conducting the word-relevance (XAI) analysis in Section 4; conducting parts of the statistical analysis, while also reviewing and validating the statistical analysis as a whole in Section 5; validating and annotating the German speeches; and co-writing the paper. 
    \item The third co-author (PA) was responsible for collecting and curating the Trump and European speech datasets; designing the coding and scoring schemes; training student annotators; authoring the sections on populism theory and the literature review; conducting inference with closed-source LLMs; formulating the research questions, hypotheses, and operationalizations presented in Section 5; interpreting the political science implications of the statistical findings; and co-writing the paper. 
\end{itemize}

\newpage

\bibliography{custom}

\begin{thebibliography}{40}
\providecommand{\natexlab}[1]{#1}

\bibitem[{Ali et~al.(2022)Ali, Schnake, Eberle, Montavon, M{\"u}ller, and Wolf}]{ali2022xai}
Ameen Ali, Thomas Schnake, Oliver Eberle, Gr{\'e}goire Montavon, Klaus-Robert M{\"u}ller, and Lior Wolf. 2022.
\newblock \href {https://proceedings.mlr.press/v162/ali22a/ali22a.pdf} {Xai for transformers: Better explanations through conservative propagation}.
\newblock In \emph{International conference on machine learning}, pages 435--451. PMLR.

\bibitem[{Aslanidis(2016)}]{aslanidis_2016}
Paris Aslanidis. 2016.
\newblock \href {https://doi.org/10.1111/1467-9248.12224} {{Is Populism an Ideology? A Refutation and a New Perspective}}.
\newblock \emph{Political Studies}, 64(1\_suppl):88--104.

\bibitem[{Aslanidis(2018)}]{aslanidis2018measuring}
Paris Aslanidis. 2018.
\newblock \href {https://www.springerprofessional.de/en/measuring-populist-discourse-with-semantic-text-analysis-an-appl/12341876} {Measuring populist discourse with semantic text analysis: an application on grassroots populist mobilization}.
\newblock \emph{Quality \& Quantity}, 52(3):1241--1263.

\bibitem[{Aslanidis(2024)}]{aslanidis-2024}
Paris Aslanidis. 2024.
\newblock \emph{Populist Mobilization}.
\newblock Oxford: Oxford University Press.

\bibitem[{Bartels(1985)}]{bartels1985resource}
Larry~M Bartels. 1985.
\newblock Resource allocation in a presidential campaign.
\newblock \emph{The Journal of Politics}, 47(3):928--936.

\bibitem[{Bommasani et~al.(2023)Bommasani, Liang, and Lee}]{bommasani2023holistic}
Rishi Bommasani, Percy Liang, and Tony Lee. 2023.
\newblock \href {https://arxiv.org/abs/2211.09110} {Holistic evaluation of language models}.
\newblock \emph{Annals of the New York Academy of Sciences}, 1525(1):140--146.

\bibitem[{Bonikowski et~al.(2022)Bonikowski, Luo, and Stuhler}]{bonikowski-2022}
Bart Bonikowski, Yuchen Luo, and Oscar Stuhler. 2022.
\newblock \href {https://doi.org/10.1177/00491241221122317} {{Politics as Usual? Measuring Populism, Nationalism, and Authoritarianism in U.S. Presidential Campaigns (1952–2020) with Neural Language Models}}.
\newblock \emph{Sociological Methods \& Research}, 51(4):1721--1787.

\bibitem[{Chiang et~al.(2024)Chiang, Zheng, Sheng, Angelopoulos, Li, Li, Zhu, Zhang, Jordan, Gonzalez et~al.}]{chiang2024chatbot}
Wei-Lin Chiang, Lianmin Zheng, Ying Sheng, Anastasios~Nikolas Angelopoulos, Tianle Li, Dacheng Li, Banghua Zhu, Hao Zhang, Michael Jordan, Joseph~E Gonzalez, and 1 others. 2024.
\newblock \href {https://arxiv.org/abs/2403.04132} {Chatbot arena: An open platform for evaluating llms by human preference}.
\newblock In \emph{Forty-first International Conference on Machine Learning}.

\bibitem[{Cortes and Vapnik(1995)}]{cortes1995support}
C.~Cortes and V.~Vapnik. 1995.
\newblock {Support Vector Networks}.
\newblock \emph{Machine Learning}, 20:273--297.

\bibitem[{Dai and Kustov(2022)}]{dai2022politicians}
Yaoyao Dai and Alexander Kustov. 2022.
\newblock {When do politicians use populist rhetoric? Populism as a campaign gamble}.
\newblock \emph{Political Communication}, 39(3):383--404.

\bibitem[{DeepSeek-AI et~al.(2025)DeepSeek-AI, Guo, Yang, Zhang, Song, Zhang, Xu, Zhu, Ma, Wang, Bi, Zhang, Yu, Wu, Wu, Gou, Shao, Li, Gao, Liu, Xue, Wang, Wu, Feng, Lu, Zhao, Deng, Zhang, Ruan, Dai, Chen, Ji, Li, Lin, Dai, Luo, Hao, Chen, Li, Zhang, Bao, Xu, Wang, Ding, Xin, Gao, Qu, Li, Guo, Li, Wang, Chen, Yuan, Qiu, Li, Cai, Ni, Liang, Chen, Dong, Hu, Gao, Guan, Huang, Yu, Wang, Zhang, Zhao, Wang, Zhang, Xu, Xia, Zhang, Zhang, Tang, Li, Wang, Li, Tian, Huang, Zhang, Wang, Chen, Du, Ge, Zhang, Pan, Wang, Chen, Jin, Chen, Lu, Zhou, Chen, Ye, Wang, Yu, Zhou, Pan, Li, Zhou, Wu, Ye, Yun, Pei, Sun, Wang, Zeng, Zhao, Liu, Liang, Gao, Yu, Zhang, Xiao, An, Liu, Wang, Chen, Nie, Cheng, Liu, Xie, Liu, Yang, Li, Su, Lin, Li, Jin, Shen, Chen, Sun, Wang, Song, Zhou, Wang, Shan, Li, Wang, Wei, Zhang, Xu, Li, Zhao, Sun, Wang, Yu, Zhang, Shi, Xiong, He, Piao, Wang, Tan, Ma, Liu, Guo, Ou, Wang, Gong, Zou, He, Xiong, Luo, You, Liu, Zhou, Zhu, Xu, Huang, Li, Zheng, Zhu, Ma, Tang, Zha, Yan, Ren, Ren, Sha, Fu, Xu, Xie, Zhang,
  Hao, Ma, Yan, Wu, Gu, Zhu, Liu, Li, Xie, Song, Pan, Huang, Xu, Zhang, and Zhang}]{deepseekai2025}
DeepSeek-AI, Daya Guo, Dejian Yang, Haowei Zhang, Junxiao Song, Ruoyu Zhang, Runxin Xu, Qihao Zhu, Shirong Ma, Peiyi Wang, Xiao Bi, Xiaokang Zhang, Xingkai Yu, Yu~Wu, Z.~F. Wu, Zhibin Gou, Zhihong Shao, Zhuoshu Li, Ziyi Gao, and 181 others. 2025.
\newblock \href {https://arxiv.org/abs/2501.12948} {Deepseek-r1: Incentivizing reasoning capability in llms via reinforcement learning}.
\newblock \emph{Preprint}, arXiv:2501.12948.

\bibitem[{Dettmers et~al.(2023)Dettmers, Pagnoni, Holtzman, and Zettlemoyer}]{dettmers2023qlora}
Tim Dettmers, Artidoro Pagnoni, Ari Holtzman, and Luke Zettlemoyer. 2023.
\newblock \href {https://arxiv.org/abs/2305.14314} {Qlora: Efficient finetuning of quantized llms}.
\newblock \emph{Advances in neural information processing systems}, 36:10088--10115.

\bibitem[{Devlin et~al.(2019)Devlin, Chang, Lee, and Toutanova}]{devlin-etal-2019-bert}
Jacob Devlin, Ming-Wei Chang, Kenton Lee, and Kristina Toutanova. 2019.
\newblock \href {https://doi.org/10.18653/v1/N19-1423} {{BERT}: Pre-training of deep bidirectional transformers for language understanding}.
\newblock In \emph{Proceedings of the 2019 Conference of the North {A}merican Chapter of the Association for Computational Linguistics: Human Language Technologies}, Minneapolis, Minnesota.

\bibitem[{Erhard et~al.(2025)Erhard, Hanke, Remer, Falenska, and Heiberger}]{Erhard_2025}
Lukas Erhard, Sara Hanke, Uwe Remer, Agnieszka Falenska, and Raphael~Heiko Heiberger. 2025.
\newblock \href {https://doi.org/10.1017/pan.2024.12} {Popbert. detecting populism and its host ideologies in the german bundestag}.
\newblock \emph{Political Analysis}, 33(1):1–17.

\bibitem[{Gemini~Team(2025)}]{gemini}
Google Gemini~Team. 2025.
\newblock \href {https://storage.googleapis.com/deepmind-media/gemini/gemini_v2_5_report.pdf} {{Gemini 2.5: Pushing the Frontier with Advanced Reasoning, Multimodality, Long Context, and Next Generation Agentic Capabilities.}}
\newblock Technical report.

\bibitem[{Gemma~Team(2025)}]{gemma_2025}
Google~DeepMind Gemma~Team. 2025.
\newblock \href {https://goo.gle/Gemma3Report} {{Gemma 3 - Technical Report}}.
\newblock Technical report.

\bibitem[{Halterman(2025)}]{Halterman_2025}
Andrew Halterman. 2025.
\newblock \href {https://doi.org/10.1017/pan.2024.31} {Synthetically generated text for supervised text analysis}.
\newblock \emph{Political Analysis}, 33(3):181–194.

\bibitem[{Hawkins et~al.(2018)Hawkins, Carlin, Littvay, and Kaltwasser}]{hawkins2018ideational}
Kirk~A Hawkins, Ryan~E Carlin, Levente Littvay, and Crist{\'o}bal~Rovira Kaltwasser. 2018.
\newblock \emph{The ideational approach to populism: Concept, theory, and analysis}.
\newblock London: Routledge.

\bibitem[{He et~al.(2023)He, Gao, and Chen}]{he2023debertav}
Pengcheng He, Jianfeng Gao, and Weizhu Chen. 2023.
\newblock \href {https://openreview.net/forum?id=sE7-XhLxHA} {De{BERT}av3: Improving de{BERT}a using {ELECTRA}-style pre-training with gradient-disentangled embedding sharing}.
\newblock In \emph{The Eleventh International Conference on Learning Representations}.

\bibitem[{Hu et~al.(2022)Hu, Shen, Wallis, Allen-Zhu, Li, Wang, Wang, and Chen}]{hu2022lora}
Edward~J Hu, Yelong Shen, Phillip Wallis, Zeyuan Allen-Zhu, Yuanzhi Li, Shean Wang, Lu~Wang, and Weizhu Chen. 2022.
\newblock \href {https://openreview.net/forum?id=nZeVKeeFYf9} {Lo{RA}: Low-rank adaptation of large language models}.
\newblock In \emph{International Conference on Learning Representations}.

\bibitem[{Klamm et~al.(2023)Klamm, Rehbein, and Ponzetto}]{klamm-etal-2023-kind}
Christopher Klamm, Ines Rehbein, and Simone~Paolo Ponzetto. 2023.
\newblock \href {https://doi.org/10.18653/v1/2023.findings-eacl.91} {{"Our kind of people? Detecting populist references in political debates"}}.
\newblock In \emph{Findings of the Association for Computational Linguistics: EACL 2023}, pages 1227--1243, Dubrovnik, Croatia. Association for Computational Linguistics.

\bibitem[{Laclau(2005)}]{laclau_2005}
Ernesto Laclau. 2005.
\newblock \emph{{On Populist Reason}}.
\newblock London: Verso.

\bibitem[{Liu et~al.(2019)Liu, Ott, Goyal, Du, Joshi, Chen, Levy, Lewis, Zettlemoyer, and Stoyanov}]{roberta-2019}
Yinhan Liu, Myle Ott, Naman Goyal, Jingfei Du, Mandar Joshi, Danqi Chen, Omer Levy, Mike Lewis, Luke Zettlemoyer, and Veselin Stoyanov. 2019.
\newblock \href {https://arxiv.org/abs/1907.11692} {Roberta: {A} robustly optimized {BERT} pretraining approach}.
\newblock \emph{CoRR}, abs/1907.11692.

\bibitem[{Llama~Team(2024)}]{llama3}
AI~@~Meta Llama~Team. 2024.
\newblock \href {https://arxiv.org/abs/2407.21783} {The llama 3 herd of models}.
\newblock \emph{Preprint}, arXiv:2407.21783.

\bibitem[{Mudde(2004)}]{Mudde_2004}
Cas Mudde. 2004.
\newblock \href {https://doi.org/10.1111/j.1477-7053.2004.00135.x} {{The Populist Zeitgeist}}.
\newblock \emph{Government and Opposition}, 39(4):541–563.

\bibitem[{Mudde(2007)}]{Mudde_2007}
Cas Mudde. 2007.
\newblock \emph{Populist Radical Right Parties in Europe}.
\newblock Cambridge: Cambridge University Press.

\bibitem[{Mudde and Kaltwasser(2017)}]{mudde2017populism}
Cas Mudde and Crist{\'o}bal~Rovira Kaltwasser. 2017.
\newblock \emph{Populism: A very short introduction}.
\newblock Oxford: Oxford University Press.

\bibitem[{Norris and Inglehart(2019)}]{Norris_Inglehart_2019}
Pippa Norris and Ronald Inglehart. 2019.
\newblock \emph{Cultural Backlash: Trump, Brexit, and Authoritarian Populism}.
\newblock Cambridge: Cambridge University Press.

\bibitem[{OpenAI(2023)}]{openai2023gpt4}
OpenAI. 2023.
\newblock \href {https://arxiv.org/abs/2303.08774} {{GPT-4 Technical Report}}.
\newblock \emph{Preprint}, arXiv:2303.08774.

\bibitem[{OpenAI~Team(2025)}]{gpt41}
OpenAI OpenAI~Team. 2025.
\newblock \href {https://openai.com/index/gpt-4-1/} {{Introducing GPT-4.1 in the API}}.
\newblock Technical report.

\bibitem[{Qwen~Team(2025)}]{qwen3technicalreport}
Qwen Qwen~Team. 2025.
\newblock \href {https://arxiv.org/abs/2505.09388} {{Qwen3 - Technical Report}}.
\newblock Technical report.

\bibitem[{Radford et~al.(2023)Radford, Kim, Xu, Brockman, Mcleavey, and Sutskever}]{pmlr-v202-radford23a}
Alec Radford, Jong~Wook Kim, Tao Xu, Greg Brockman, Christine Mcleavey, and Ilya Sutskever. 2023.
\newblock \href {https://proceedings.mlr.press/v202/radford23a.html} {Robust speech recognition via large-scale weak supervision}.
\newblock In \emph{Proceedings of the 40th International Conference on Machine Learning}, volume 202 of \emph{Proceedings of Machine Learning Research}, pages 28492--28518. PMLR.

\bibitem[{Shaw et~al.(2024)Shaw, Althaus, and Panagopoulos}]{shaw2024battleground}
Daron~R Shaw, Scott~L Althaus, and Costas Panagopoulos. 2024.
\newblock \emph{Battleground: Electoral College Strategies, Execution, and Impact in the Modern Era}.
\newblock Oxford University Press.

\bibitem[{Stavrakakis(2024)}]{stavrakakis-2024}
Yannis Stavrakakis. 2024.
\newblock \href {https://www.routledge.com/Populist-Discourse-Recasting-Populism-Research/Stavrakakis/p/book/9781032284927} {\emph{Populist Discourse: Recasting Populism Research}}.
\newblock London: Routledge.

\bibitem[{Stavrakakis and Katsambekis(2024)}]{stavrakakis2024research}
Yannis Stavrakakis and Giorgos Katsambekis. 2024.
\newblock \emph{Research Handbook on Populism}.
\newblock Cheltenham: Edward Elgar Publishing.

\bibitem[{Tao et~al.(2025)Tao, Zhan, Zhou, Kang, and Sun}]{tao2025measuring}
Yuzhou Tao, Zhiqin Zhan, Han Zhou, Jingshi Kang, and Shaojing Sun. 2025.
\newblock \href {https://www.tandfonline.com/doi/full/10.1080/17544750.2024.2420622} {Measuring chinese online populist discourse: an automated semantic text analysis method}.
\newblock \emph{Chinese Journal of Communication}, 18(2):121--141.

\bibitem[{Wei et~al.(2022)Wei, Wang, Schuurmans, Bosma, Ichter, Xia, Chi, Le, and Zhou}]{wei-et-al-2022}
Jason Wei, Xuezhi Wang, Dale Schuurmans, Maarten Bosma, Brian Ichter, Fei Xia, Ed~H. Chi, Quoc~V. Le, and Denny Zhou. 2022.
\newblock \href {https://dl.acm.org/doi/10.5555/3600270.3602070} {Chain-of-thought prompting elicits reasoning in large language models}.
\newblock In \emph{Proceedings of the 36th International Conference on Neural Information Processing Systems}, NeurIPS '22, Red Hook, NY, USA.

\bibitem[{Weyland(2001)}]{weyland_2001}
Kurt Weyland. 2001.
\newblock \href {https://doi.org/10.2307/422412} {{Clarifying a Contested Concept: Populism in the Study of Latin American Politics}}.
\newblock \emph{Comparative Politics}, 34(1):1--22.

\bibitem[{Weyland(2017)}]{weyland_2017}
Kurt Weyland. 2017.
\newblock \href {https://doi.org/10.1093/oxfordhb/9780198803560.013.3} {{Populism: A Political-Strategic Approach}}.
\newblock In Cristóbal~Rovira Kaltwasser, Paul Taggart, Paulina~Ochoa Espejo, and Pierre Ostiguy, editors, \emph{The Oxford Handbook of Populism}, pages 48--73. Oxford University Press, Oxford.

\bibitem[{Zhang and Schroeder(2024)}]{zhang_2024}
Yuan Zhang and Ralph Schroeder. 2024.
\newblock \href {https://doi.org/10.1177/20563051241229659} {{“It’s All About US vs THEM!”: Comparing Chinese Populist Discourses on Weibo and Twitter}}.
\newblock \emph{Social Media + Society}, 10(1):20563051241229659.

\end{thebibliography}

\appendix
\section{Additional Experimental Details}
\label{sec:exp_details}

\subsection{Baseline}
\label{sec:baseline_details}

We train a linear SVM-based classifier using the Scikit-learn library.\footnote{\url{https://scikit-learn.org/stable/}} We use the \texttt{TfidfVectorizer} to vectorize the inputs with min\_df=20, max\_df=0.5, for 10K $n$-grams, where $n\in[1,2,3]$ with a lower-cased vocabulary. We use the \texttt{svm.SVC} with a linear kernel as a multi-output classifier. 

\subsection{PLMs}
\label{sec:plm_details}

\paragraph{Fine-tuning SetUp} We fine-tune both PLMs (RoBERTa and DeBERTa-V3) as 3-way multi-label classifiers using a shallow MLP on top of the final \texttt{<s>}, also known as \texttt{[CLS]}, representation. We train both models for a total of 5 epochs (iterations over the training examples), using cosine learning rate scheduling with a maximum learning rate of $1e\!-\!5$, a warmup ratio of 0.1, and weight decay of 0.01.\footnote{We select the core hyper-parameters after a brief grid-search for learning rate ($lr\in[1e-5, 3e-5, 5e-5]$ and epochs ($e\in[3,5,10]$) using a subset of the training set as a development set, tracking the macro-F1.} We use mini-batches of 32 examples, the maximum we can use, to maximize the potential of having populist (labeled as PC and/or AE) examples in every mini-batch,  since positive examples are very sparse (7.4\% of the training set), and even sparser considering individual categories (AE, PC). We also truncate all examples to 64 tokens, which is much larger compared to the average sentence length (approx. 15 words), to speed up training.

\paragraph{Upsampling} To further tackle the intense class-imbalance, we use a trivial up-sampling (5$\times$) of the positive examples (ending up with approx. 37\% positive examples) for fine-tuning, which leads to improved performance compared to using the original dataset (no augmentation, see Table~\ref{tab:roberta_aug}). In Appendix~\ref{sec:augmentation}, we present results using more elaborate data augmentation strategies with negative results (Appendix~\ref{sec:augmentation}).

\begin{table}[t]
    \centering
    \begin{tabular}{l|c|c|c|c}
         \textbf{Augmentation} & \textbf{N} & \textbf{PC} & \textbf{AE} & \textbf{Avg} \\
         \midrule
         None & .975 & .625 & .566 & .722 \\
         \midrule
         Up-sampling (5$\times$) & .975 & .661 & .602 & .746 \\
         Synthetic & .969 & .589 & .574 & .710 \\
         Curated Synthetic & .975	& .690	& .584	& .750 \\
         Paraphrased & .972	& .629	& .586	& .729 \\
    \end{tabular}
    \caption{RoBERTa results using different augmentation techniques (settings).}
    \label{tab:roberta_aug}
\end{table}

\subsection{LLMs}

\paragraph{Prompts and SetUp} In Table~\ref{tab:prompts}, we present the exact (verbatim) wording of all examined prompts. As mentioned above, the vast majority of our experiments are with LLMs out-of-the-box. We use 4-bit quantized versions. We run 6 runs for all open-weight models using 3 different seeds ($s\in[21,42,84]$) for all random engines, e.g., numpy, torch, etc. At the same time, we also present the options in two different orders: (i) where a-d are neutral, anti-elitism, people-centrism in order, both, and (ii) where a-d are \textit{both}, \textit{anti-elitism}, \textit{people-centrism}, and \textit{neutral} in order. We report the mean and standard deviation per class across all 6 runs.

\paragraph{Finetuned LLMs}We additionally present results with fine-tuned versions of Llama 3.1 (8B), and Qwen (8B) using LoRA adapter ($\alpha=32$, $r=16$, and a dropout rate of 0.05) targeting the projection layers, following best practices. We fine-tune these models for 1 epoch, since they rapidly overfit, as the main point of fine-tuning is to optimize the selection among 4 predefined options (a-d). During training, we mask the whole input (instruction) up to the expected generated when computing the loss.
 We do not run different option orders for fine-tuned models, since they have been fine-tuned to follow a pre-defined order.

\section{Additional Experiments}
\label{sec:add_experiments}

\subsection{Data Augmentation}
\label{sec:augmentation}

We consider alternative data augmentation strategies to improve PLMs' performance. All strategies aim at augmenting the training set with positive (labeled as PC and/or AE) examples. The strategies we examine are as follows: (a) \textbf{Up-sampling (5$\times$)}: Up-sampling the ratio of examples labeled with positive categories (AE and/or PC) by a factor of 5, resulting in 5.6K positive examples in total, (b) \textbf{Synthetic}: Augmentation with 1.2K synthetic positive (AE and/or PC) examples generated by Llama 3.1 (70B) and ChatGPT.
(c) \textbf{Curated Synthetic}: Augmentation with a manually curated subset of 500 examples selected from the 1.2K synthetic examples.
(d) \textbf{Paraphrased}: Augmentation with 3.5K examples labeled with positive categories (AE and/or PC) paraphrased by Llama 3.1 (8B).

As shown in Table~\ref{tab:roberta_aug}, naive up-sampling (5$\times$) leads to a substantial performance improvement compared to the baseline (\emph{None}), which uses the original unbalanced dataset. By contrast, the vanilla generation of synthetic data (\emph{Synthetic}) has a negative impact. Manually curating (sub-sampling) the synthetic data (\emph{Curated Synthetic}) performs comparably to naive up-sampling. Lastly, augmentation via paraphrasing (\emph{Paraphrased}) has a trivial impact on performance.

\begin{table}[t]
    \centering
    \begin{tabular}{l|c|c|c|c}
         \textbf{Setting} & \textbf{N} & \textbf{PC} & \textbf{AE} & \textbf{Avg} \\
         \midrule
         Baseline & .975 & .661 & .602 & .746 \\
         Context-Aware & .975 & .649 & .575 & .733 \\
    \end{tabular}
    \caption{Results of RoBERTa with and without context.}
    \label{tab:hier-roberta}
\end{table}

\subsection{Context-Aware RoBERTa}
\label{sec:hier_roberta}

We also considered a variant of RoBERTa that is enhanced with additional context. Specifically, we use an input, the examined sentence, as before, and up to 7 preceding sentences separated by the special \texttt{<sep>} token. The hypothesis is that since populist invocations are heavily context-dependent, presenting context alongside the sentence will lead to performance improvement. As we observe in the results of Table~\ref{tab:hier-roberta}, the performance deteriorates. We hypothesize that RoBERTa cannot effectively use the context as additional information and is negatively biased by the context, i.e., if a populist sentence is preceding then the model classifies the examined sentence as such, similar to Qwen 3 (14B) prompted in a similar setting (Table~\ref{tab:llm_modes} in Section~\ref{sec:prompt_tuning}). The context-aware model is also substantially slower encoding 512 (8$\times$64) tokens in comparison to 64 tokens.

 \begin{table}[t]
    \centering
    \resizebox{\columnwidth}{!}{
    \begin{tabular}{l|c|c|c|c}
         \textbf{Prompt Setting} & \textbf{N} & \textbf{AE} & \textbf{PC} & \textbf{Avg} \\
         \midrule
         Baseline & .838 & .341 & .252 & .477 \\
         \midrule
         Context-Aware & .507 & .212 & .116 & .278 \\
         Distribution-Aware & \textbf{.879} & \textbf{.368} & \textbf{.269} & \textbf{.505} \\ 
         K-Shot (K=8) & .543 & .201 & .199 & .314 \\
         K-Shot (K=32) & .704 & .275 & .186 & .388 \\
    \end{tabular}
    }
     \vspace{-1mm}
    \caption{Results of Llama 3.1 (8B) performance across different prompt settings.}
    \vspace{-3mm}
    \label{tab:llm_modes_2}
\end{table}

\subsection{Llama 3.1 - Prompt Tuning}
\label{sec:add_experiments_llama}

Table~\ref{tab:llm_modes_2} presents results for Llama 3.1 (8B) across different prompting strategies, following a similar analysis to that of Qwen 3 (14B) in Section~\ref{sec:llm_analysis}. As with Qwen 3 (14B), we observe that providing preceding sentences as context (\emph{Context-Aware}) harms performance. In contrast, there is a considerable performance improvement (approx.~-3\%) when we present the expected label distribution (\emph{Distribution-Aware}) compared to our base prompt (\emph{Base}). Lastly, we observe that presenting labeled examples (\emph{K-Shot}) has a substantially negative impact (approx. -10-15\%) in the case of Llama 3.1, which may indicate that smaller LLMs may struggle to effectively utilize few-shot demonstrations. By comparison, in Section~\ref{sec:llm_analysis}, we find that both Qwen 3 (14B) and Gemini Flash 2.5 do benefit from few-shot demonstrations (see Tables~\ref{tab:llm_modes}-\ref{tab:gemini_modes}).

\begin{table}[t]
    \centering
    \small
    \begin{tabular}{r|c|c|c}
         \multirow{2}{4.5em}{\textbf{Speaker}} & \multicolumn{3}{c}{\textbf{Models}} \\
         &  RoBERTa  & Qwen 3 & Gemini 2.5 \\
         \midrule
         \multicolumn{4}{c}{In-Domain}\\
         \midrule
          Trump (US) & \textbf{.746} & .602 & .620\\
         \midrule
        \multicolumn{4}{c}{Out-Of-Domain (Translated)}\\
        \midrule
         LePen (FR) & .487 & .609 & \textbf{.650} \\
         Weidel (DE) & .329 & \textbf{.437} & .314 \\
         Zemmour (FR) & .388 & .583 & \textbf{.554} \\
         Kickl (AU) & .408 & .473 & \textbf{.558}\\
         Tsipras (GR) & .550 & \textbf{.658} & .586 \\
         \midrule
         Overall & .508 & \textbf{.613} & .590 \\
         \midrule
        \multicolumn{4}{c}{Out-Of-Domain (Original)}\\
        \midrule
         LePen (FR) & - & .571 & \textbf{.685} \\
         Weidel (DE) & - & \textbf{.479} & .427 \\
         Zemmour (FR) & - & \textbf{.541} & .522 \\
         Kickl (AU) & - & .507 & \textbf{.516} \\
         Tsipras (GR) & - & .629  & \textbf{.634} \\
         \midrule
         Overall & - & \textbf{.607} & .606 \\
         
    \end{tabular}
    \caption{Performance of the best-performing examined models on the \textsc{EU-OOD} dataset. -EN refers to the translated version, and -ML to the original multi-lingual version, including German, French, and Greek.}
    \label{tab:eu_speeches_extra}
    \vspace{-3mm}
\end{table}

\subsection{EU-OOD Results per Speaker}
\label{sec:per_eu_speaker}

In Table~\ref{tab:eu_speeches_extra}, we present breakdown results per speaker on the EU-OOD dataset. Given the limited number of samples, i.e.,  one speech with a few hundred sentences per speaker, we believe that no meaningful analysis can be substantiated considering the models' performance. In Table~\ref{tab:eu_speeches_percentage}, we present the populist ratio, i.e. the percentage of populist (\textit{anti-elitism} and/or \textit{people-centrism}) sentences among all sentences, per speaker. As mentioned in Section~\ref{sec:eu_benchmarking}, in some cases, e.g., Alice Weidel and Herbert Kickl, the populist content is very low, raising doubts on whether these politicians are properly designated as populists in the academic literature; their misclassification as populists seems to be mainly driven by rhetoric related to the concepts of nationalism and nativism. Further research is needed to substantiate this claim by carefully examining larger corpora.

\begin{table}[t]
    \centering
    \small
    \begin{tabular}{r|r|r}
         \textbf{Speaker Name} & Students & Gemini 2.5 Flash \\
         \midrule
          Trump (US) & 7.0\% & 20.2\% \\
          \midrule
         LePen (FR) & 13.2\% & 37.8\% \\
         Weidel (DE) & 1.2\% & 9.9\% \\
         Zemmour (FR) & 9.6\% & 22.8\% \\
         Kickl (AU) & 3.3\% & 15.9\% \\
         Tsipras (GR) & 26.8\% & 48.3\% \\
         \midrule
         Overall &  12.9\% & 26.6\% \\
    \end{tabular}
    \caption{Populism ratio per speaker on the \textsc{EU-OOD} dataset based on the manual coding of the students or the predictions of Gemini 2.5 Flash.}
    \label{tab:eu_speeches_percentage}
    \vspace{-3mm}
\end{table}

\section{Annotation Process}
\label{sec:guidelines}

Four undergraduate students from Yale University's Department of Political Science were recruited to help build the annotated dataset of Trump speeches. They received extensive training from our domain expert, using a detailed coding scheme (see  ``Coding Manual'' below) designed to identify key markers of populist discourse and to distinguish populism from related political science concepts such as nationalism, nativism, and socialism.

The training phase began with a pilot stage in which the students annotated out-of-domain sample speeches and received detailed feedback. This was followed by a second stage in which the students independently annotated a subset of the Trump speeches drawn from the target dataset, without further feedback. On this in-domain material, they achieved an inter-annotator agreement of Krippendorff’s $\alpha$ = 0.751.

After establishing sufficient agreement, the remaining speeches were divided among the students for independent annotation. This final phase produced a dataset of 15,025 sentences, with an average sentence length of approximately 15 words.

Overall, the same procedure was used for coding the European annotated dataset. Students were recruited from Panteion University of Social and Political Sciences (Athens, Greece) to code the French and Greek speeches. Speeches by Kickl and Weidel were annotated by one of the co-authors who is fluent in German, under the supervision of our domain expert.

\begin{table*}[]
    \centering
    \resizebox{\textwidth}{!}{
    \begin{tabular}{r|p{6cm}|p{6cm}|p{6cm}}
         \textbf{POS} & \textbf{Neutral} & \textbf{Anti-elitism} & \textbf{People-centrism}  \\
         \midrule
         ADJ & great, good, right, bad, fantastic & rigged, corrupt, entire, political, special & corrupt, rigged, broken, political, lose\\ \midrule
         NOUN & people, percent, numbers, job, deal & system, establishment, people, government, swamp & people, government, power, interests, system\\ \midrule
         PRON & you, i, that, they, it & the, our, you, ourselves, us & every, you, ourselves, the, our \\ \midrule
         PROPN & trump, obamacare, open, hampshire, mr.& washington, hillary, serve, american, d.c. & american, serve, washington, americans, solve\\ \midrule
         VERB & know, look, doing, do, see & win, hear, organized, believes, believe & believe, believes, trying, believing, ruled\\
    \end{tabular}
    }
    \vspace{-2mm}
    \caption{Top-5 relevant tokens per class across the most common Part-Of-Speech (POS) tags, applied to correctly classified sentences for the fine-tuned RoBERTa model. Relevance scores were computed with LRP.}
    \vspace{-3mm}
    \label{tab:xai}
\end{table*}

\paragraph{Coding Manual}

The coding manual outlines a structured, step-by-step process for annotators to identify and classify populist discourse in political texts.\footnote{Available in PDF format \href{https://drive.google.com/file/d/1T0p9jl3RNfNfr_egbc4bOQ6IABmreW8w/view?usp=sharing}{here}.} Annotators are instructed to begin with a holistic reading of the entire speech to grasp its overall context and to mentally map out the in-group (``we'') and out-group (``they'') actors. Coding is done at the sentence level, with each sentence assessed for the presence of social actors (the ``actor set''). The first step is to check whether the sentence contains any actors; if not, it is skipped. If actors are present, the coder must determine whether the sentence reflects \textit{people-centrism}, i.e. references to a moral and sovereign ``people'', and/or \textit{anti-elitism}, i.e. references to a corrupt elite usurping political power. These two codes can appear independently or together, depending on the content of the sentence. Coders are trained to assess both explicit and, where justified, implicit references, and to use context to resolve ambiguities. A flowchart is provided to assist in decision-making, but consistency and neutrality in interpretation are emphasized throughout.

\begin{table*}
\centering
\begin{tabular}{lcccc}
\hline
\textbf{Comparison} & \textbf{t-value} & \textbf{p-value} & \textbf{Mean Difference ($\Delta$)} & \textbf{Cohen's $d$} \\
\hline
2016 Primaries vs 2016 Campaign & -11.867 & .000$^{***}$ & -4.63 & -1.303 \\
2016 Primaries vs 2020 Campaign & -2.816  & .005$^{**}$  & -0.78 & -0.342 \\
2016 Primaries vs 2024 Campaign & .777   & .438 (n.s.)  & +0.16 & .084  \\
2016 Campaign vs 2020 Campaign  & 7.315   & .000$^{***}$ & +3.86 & .974  \\
2016 Campaign vs 2024 Campaign  & 11.307  & .000$^{***}$ & +4.80 & 1.367  \\
2020 Campaign vs 2024 Campaign  & 4.728   & .000$^{***}$ & +0.94 & .622  \\
\hline
\end{tabular}
\caption{Pairwise t-tests comparing $\mathrm{PDI}$ scores across campaign periods, with Cohen’s $d$ effect sizes.}
\label{tab:pairwise_ttests_pdi}
\end{table*}

\begin{table*}
\centering
\begin{tabular}{lcccc}
\hline
\textbf{Comparison} & \textbf{t-value} & \textbf{p-value} & \textbf{Mean Difference ($\Delta$)} & \textbf{Cohen's $d$} \\
\hline
2016 Primaries vs 2016 Campaign & -12.088 & .000$^{***}$ & -6.49 & -1.328 \\
2016 Primaries vs 2020 Campaign & -3.347  & .001$^{***}$ & -1.21 & -0.406 \\
2016 Primaries vs 2024 Campaign & .575   & .565 (n.s.)  & +0.16 & .062  \\
2016 Campaign vs 2020 Campaign  & 7.102   & .000$^{***}$ & +5.28 & .945  \\
2016 Campaign vs 2024 Campaign  & 11.090  & .000$^{***}$ & +6.65 & 1.340  \\
2020 Campaign vs 2024 Campaign  & 4.822   & .000$^{***}$ & +1.37 & .634  \\
\hline
\end{tabular}
\caption{Pairwise t-tests comparing $\mathrm{WPDI}$ scores across campaign periods, with Cohen’s $d$ effect sizes.}
\label{tab:pairwise_ttests_wpdi}
\end{table*}

\begin{table*}
\centering
\resizebox{\textwidth}{!}{
\begin{tabular}{c|l|ccc|ccc}
\multirow{2}{*}{\textbf{Campaign}} & \multirow{2}{*}{\textbf{Clustering Scheme}} & \multicolumn{3}{c|}{\textbf{PDI}} & \multicolumn{3}{c}{\textbf{WPDI}} \\
& & \textbf{Difference} & \textbf{p-value} & \textbf{Cohen's d} & \textbf{Difference} & \textbf{p-value} & \textbf{Cohen's d} \\
\midrule
\textbf{2016} & Ballotpedia & -0.445 & .785 & -0.090 & -0.114 & .960 & -0.016 \\
& High Attention & -1.170 & .182 & -0.238 & -1.414 & .253 & -0.204 \\
\midrule
\textbf{2020} & Ballotpedia & -0.584 & .290 & -0.291 & -0.547 & .482 & -0.193 \\
& High Attention & -0.214 & .629 & -0.106 & -0.039 & .951 & -0.014 \\
\midrule
\textbf{2024} & Ballotpedia & -0.577 & .001*** & -0.579 & -0.863 & .001*** & -0.588 \\
& High Attention & -0.200 & .270 & -0.193 & -0.223 & .403 & -0.146 \\
\bottomrule
\end{tabular}
}
\caption{Statistical Test Results -- Swing ($S$) vs Non-Swing ($\neg S$) State Populist Discourse. Difference is defined as: $\overline{\mathrm{PDI}_S} - \overline{\mathrm{PDI}_{\neg S}}$. Negative values (-) indicate lower populist rhetoric in swing states. *** p < 0.001.}
\label{tab:q2_ttest_results}
\end{table*}

\begin{table*}
\centering
\resizebox{\textwidth}{!}{
\begin{tabular}{l|c|cc|cc|cc}
\toprule
\multirow{2}{*}{\textbf{Metric}} & \multirow{2}{*}{\textbf{Clustering Scheme}} & \multicolumn{2}{c|}{\textbf{2016 Campaign}} & \multicolumn{2}{c|}{\textbf{2020 Campaign}} & \multicolumn{2}{c}{\textbf{2024 Campaign}} \\
& & \textbf{Swing} & \textbf{Non-Swing} & \textbf{Swing} & \textbf{Non-Swing} & \textbf{Swing} & \textbf{Non-Swing} \\
\midrule
\multirow{4}{*}{\textbf{PDI}} & \multirow{2}{*}{Ballotpedia} & 6.269 & 6.714 & 2.350 & 2.933 & 1.226 & 1.804 \\
& & (n=123) & (n=10) & (n=82) & (n=16) & (n=73) & (n=68) \\
& \multirow{2}{*}{High Attention} & 5.845 & 7.015 & 2.379 & 2.594 & 1.431 & 1.631 \\
& & (n=81) & (n=52) & (n=68) & (n=30) & (n=89) & (n=52) \\
\midrule
\multirow{4}{*}{\textbf{WPDI}} & \multirow{2}{*}{Ballotpedia} & 8.662 & 8.776 & 3.302 & 3.850 & 1.607 & 2.470 \\
& & (n=123) & (n=10) & (n=82) & (n=16) & (n=73) & (n=68) \\
& \multirow{2}{*}{High Attention} & 8.118 & 9.531 & 3.380 & 3.419 & 1.941 & 2.165 \\
& & (n=81) & (n=52) & (n=68) & (n=30) & (n=89) & (n=52) \\
\bottomrule
\end{tabular}
}
\caption{Descriptive Statistics: Mean $\mathrm{PDI}$ and WPDI scores by Campaign, State Type, and Clustering Scheme. Sample sizes (n) in parentheses.}
\label{tab:q2_descriptive_stats}
\end{table*}

\begin{table*}
\centering
\begin{tabular}{llrcr}
\hline
\textbf{Populism-type} & \textbf{Comparison} & \textbf{t-value} & \textbf{p-value} & \textbf{Cohen's d}  \\
\hline
\multirow{3}{*}{Overall}& Opening vs Closing & - 5.357 & .000$^{***}$ & -0.318\\
& Opening vs Body & 2.677 & .023$^{*}$ & 0.169\\
& Body vs Closing & -8.851 & .000$^{***}$ & -0.565\\ \hline
\multirow{3}{*}{Anti-elitism}& Opening vs Closing & 0.765 & 1.0 (n.s.)& 0.045\\
& Opening vs Body & 1.623 & .315 (n.s.) & 0.266\\
& Body vs Closing & -0.765 & 1.0 (n.s.) & -0.048\\ \hline
\multirow{3}{*}{People-centrism}& Opening vs Closing & -10.910 & .000$^{***}$ & -0.658\\
& Opening vs Body & 4.479 & .000$^{***}$ & 0.266 \\
& Body vs Closing & -16.703 & .000$^{***}$ & -1.016\\
\hline
\end{tabular}
\caption{Pairwise t-tests comparing $\mathrm{PDI}$ scores across \textit{Opening ($20\%$)}, \textit{Body ($60\%$)} and \textit{Closing ($20\%$)} bins for all three populism types.}
\label{tab:opening_closing_analysis}
\end{table*}

\section{Analysis of Donald Trump's Populism - Additional Results}
\label{sec:trump_analysis_add}

\begin{figure*}
    \centering
    \resizebox{\textwidth}{!}{
    \includegraphics{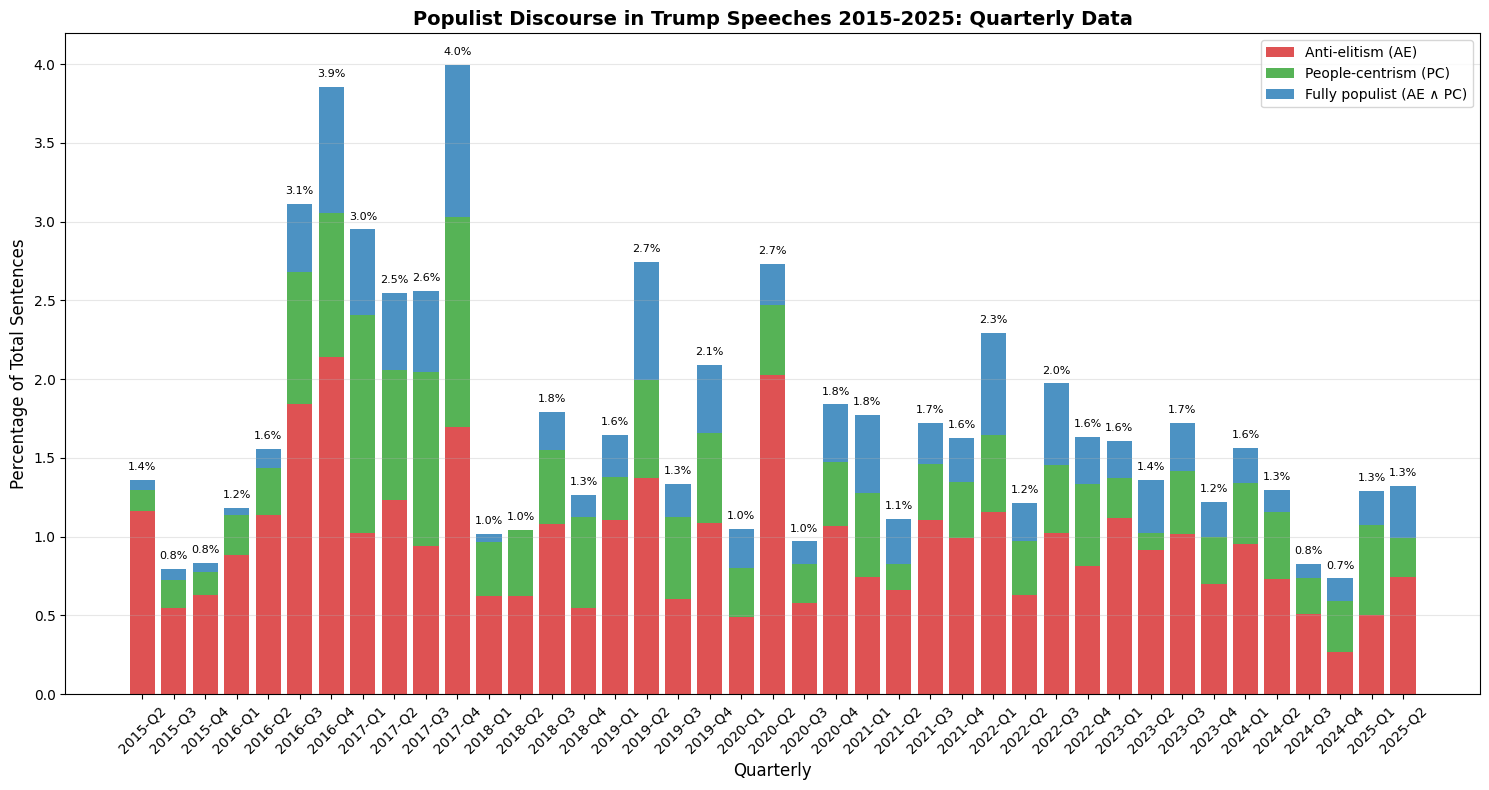}
    }
    \caption{Distribution of populist sentences among different types (anti-elitism, people-centrism, or both) across yearly quarters.}
    \label{fig:bar-chart}
\end{figure*}

\begin{figure*}
\centering
    \includegraphics[width=\textwidth]{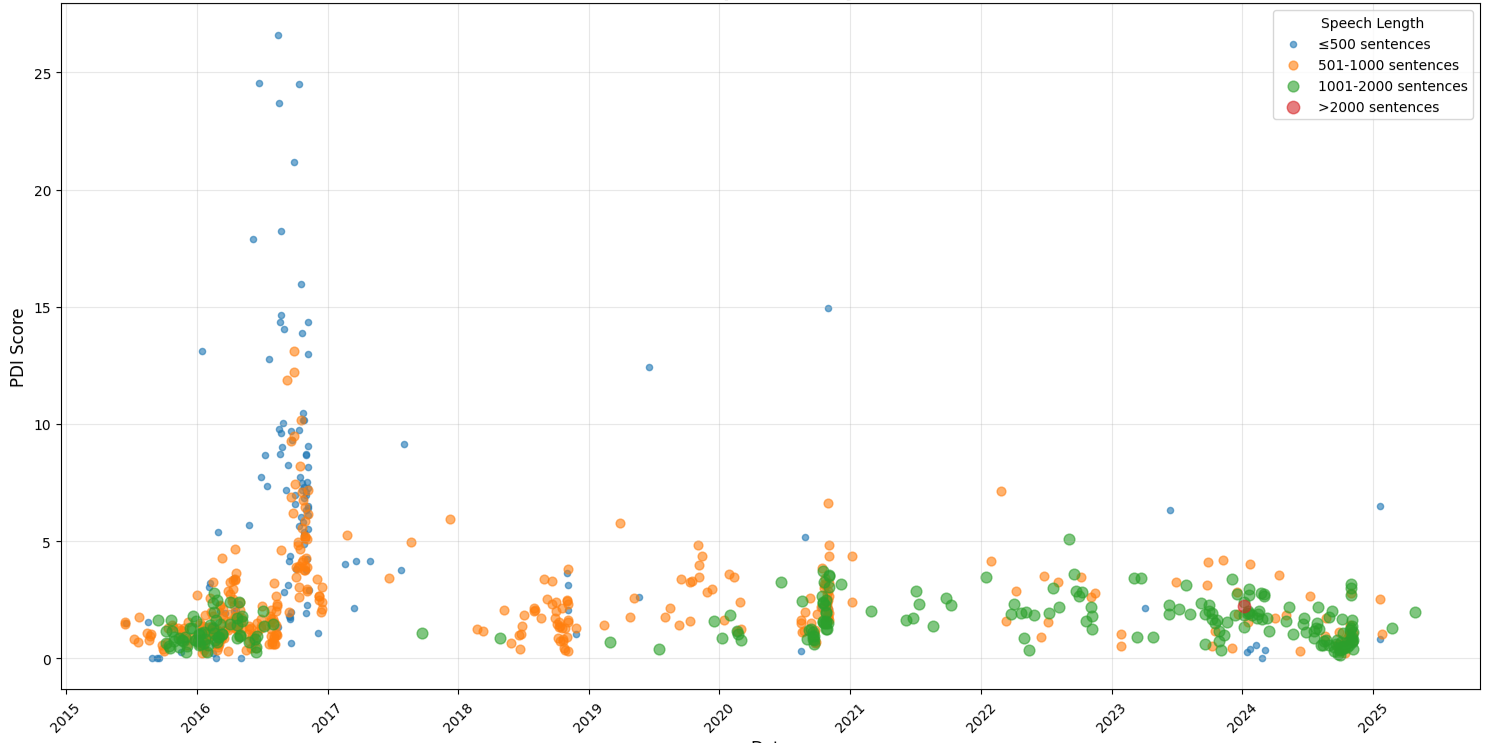}
    \caption{PDI scores per speech, June 16, 2015 to April 29, 2025.}
    \label{fig:figure2}
\end{figure*}

\begin{figure*}
    \centering
    \includegraphics[width=1\textwidth]{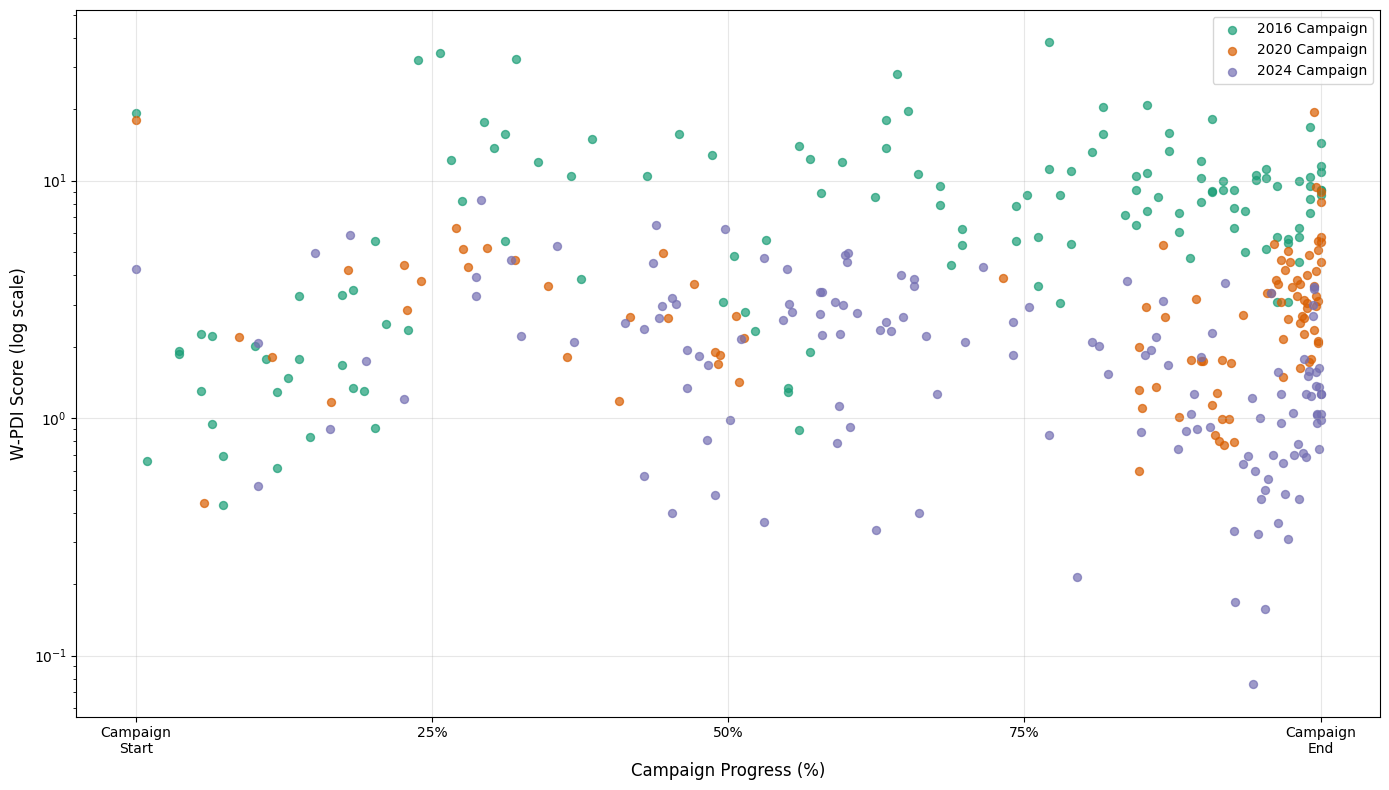}
    \caption{WPDI across presidential campaigns, aligned for relative progress.}
    \label{fig:enter-label}
\end{figure*}

\begin{figure*}
    \centering
    \resizebox{\textwidth}{!}{
    \includegraphics{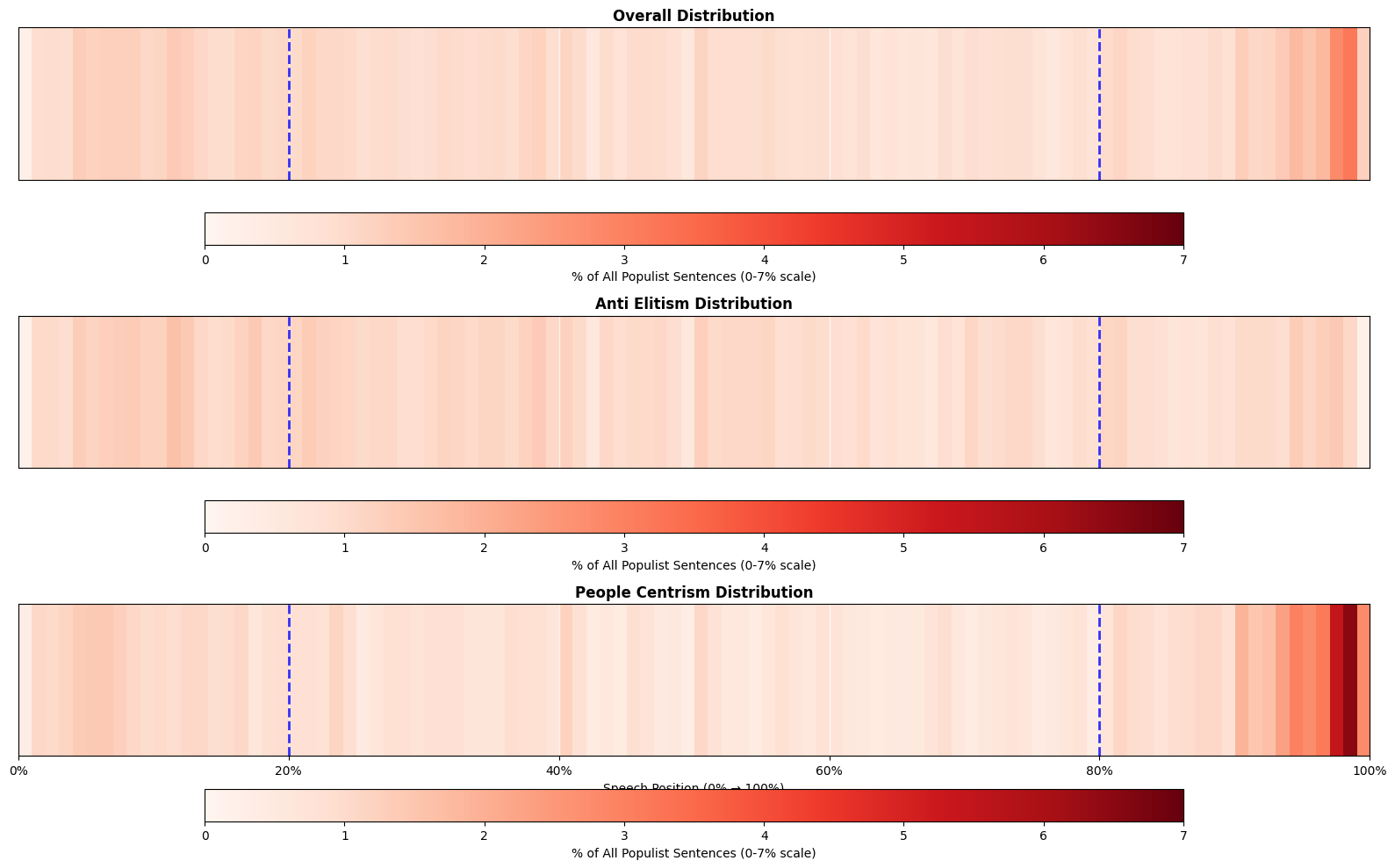}
    }
    \caption{Continuous Distribution of Populist Discourse Across Speech Positions.}
    \label{fig:heatstrips}
\end{figure*}

In Table~\ref{tab:pairwise_ttests_pdi}, we present detailed results of the pairwise t-tests comparing $\mathrm{PDI}$ scores across campaign periods (RQ3 - Section~\ref{sec:trump_analysis}). In Table~\ref{tab:swing_states}, we present the clustering of swing states vs non-swing states by campaign period. In Table~\ref{tab:q2_ttest_results}, we present the statistical test results for the swing states vs non-swing states analysis (RQ2 - Section~\ref{sec:trump_analysis}). In Figure~\ref{fig:speech_5_binsenter-label}, we present the distribution of populist content across speech related to RQ3 in Section~\ref{sec:trump_analysis}. Figures~\ref{fig:bar-chart}-\ref{fig:heatstrips} depict additional visualization of the results.

\begin{table*}[t]
    \centering
    \small
    \begin{tabular}{c|p{12cm}}
         \textbf{Setting} & \textbf{Prompt Text} \\
         \midrule
         \emph{Base}  & You are a helpful AI assistant with expertise in identifying populism in public discourse. Populism can be defined as an anti-elite discourse in the name of the "people". In other words, populism emphasizes the idea of the common "people" and often positions this group in opposition to a perceived elite group.\\
         \\
         & There are two core elements in identifying populism: (i) anti-elitism, i.e., negative invocations of "elites", and (ii) people-centrism, i.e., positive invocations of the "people". \\
         \\
         & You must classify each sentence in one of the following categories: \\ 
         \\
         & (a) No populism. \\
         & (b) Anti-elitism, i.e., negative invocations of "elites". \\
         & (c) People-centrism, i.e., positive invocations of the "People". \\
         & (d) Both people-centrism and anti-elitism populism.\\
         \midrule
         \emph{Content-Aware} & \emph{[Base]} + \\
         \\
         & Here are the preceding sentences for context: \\
         & [Up to 5 preceding sentences] \\
         \\
         & When classifying a sentence, focus primarily on the content of that specific sentence. Use the context of preceding sentences only to resolve coreferences (e.g., identifying who "they" or "you" refer to) or to disambiguate when the sentence is ambiguous on its own. \\
         \midrule
         \emph{Distribution-Aware} & \emph{[Base]} + \\
         \\
         & The label distribution is (a) No populism (92\%), (b) Anti-elitism (4\%), (c) People-centrism (2\%), (d) Both people-centrism and anti-elitism (2\%). \\
         \midrule
         \emph{K-shot} & \emph{[Base]} +\\
         \\
         & The following sentences are in category (a) No populism: \\
         & [K/4 examples] \\
         & The following sentences are in category (b) Anti-elitism populism: \\
         & [K/4 examples] \\
         & The following sentences are in category (c) People-centrism populism: \\
         & [K/4 examples] \\
         & The following sentences are in category (d) Both people-centrism and anti-elitism populism: \\
         & [K/4 examples] \\
         \midrule
         \emph{RAG-Shot} & \emph{[Base]} +\\
         \\
         & Here are the most similar [K] sentences from the training set, accompanied by their label: \\
         & [K most-similar examples with labels] \\
         \\
         & When classifying a sentence, focus primarily on the content of that specific sentence. \\
    \end{tabular}
    \caption{Prompt Settings explored in Section~\ref{sec:results} - ``Qwen 3 Exploration'' and ``Gemini 2.5 Exploration''. All presented prompt settings are followed by the question: \emph{``Which is the most relevant category for the sentence: $S$?''}}
    \label{tab:prompts}
\end{table*}

\end{document}